\documentclass[pdflatex,sn-mathphys-num]{sn-jnl}

\usepackage{graphicx}%
\usepackage{multirow}%
\usepackage{amsmath,amssymb,amsfonts}%
\usepackage{amsthm}%
\usepackage{mathrsfs}%
\usepackage[title]{appendix}%
\usepackage{textcomp}%
\usepackage{manyfoot}%
\usepackage{booktabs}%
\usepackage{algorithm}%
\usepackage{algorithmicx}%
\usepackage{caption}
\usepackage{geometry}
\usepackage{hyperref}
\usepackage[table,xcdraw]{xcolor}
\usepackage{float}       
\usepackage{adjustbox} 
\usepackage{tabularx}
\usepackage{longtable}
\usepackage{lmodern} 
\usepackage{tcolorbox}

\theoremstyle{thmstyleone}%
\theoremstyle{thmstyletwo}%

\theoremstyle{thmstylethree}%
\newcommand{\correct}[2]{\colorbox{green!25}{\includegraphics[height=0.3cm]{#1}~#2}}
\newcommand{\incorrect}[2]{\colorbox{red!25}{\includegraphics[height=0.3cm]{#1}~#2}}
\newcommand{\bestmodel}[1]{\underline{#1}}   
\newcommand{\besttype}[1]{\textbf{#1}}       

\begin{document}

\title{From Words to Proverbs: Evaluating LLMs’ Linguistic and Cultural Competence in Saudi Dialects with Absher}


 \author[1, 2]{\fnm{Renad} \sur{Al-Monef}}\email{renadsalem01@gmail.com}

 \author[3]{\fnm{Hassan} \sur{Alhuzali}}\email{hrhuzali@uqu.edu.sa}

 \author[4]{\fnm{Nora} \sur{Alturayeif}}\email{nsalturayeif@iau.edu.sa}

 \author*[1, 2]{\fnm{Ashwag} \sur{Alasmari}}\email{aasmry@kku.edu.sa}

 \affil*[1]{\orgname{Department of Computer Science, King Khalid University}, \city{Abha}, \country{Saudi Arabia}}
 \affil[2]{\orgdiv{Center for Artificial Intelligence}, \orgname{King Khalid University}, \orgaddress{\city{Abha}, \country{Saudi Arabia}}}

 \affil[3]{\orgname{Department of Computer Science and Artificial Intelligence, Umm Al-Qura University}, \city{Makkah}, \country{Saudi Arabia}}

 \affil[4]{\orgname{Imam Abdulrahman Bin Faisal University}, \city{Dammam}, \country{Saudi Arabia}}


\keywords{Large Language Models, Natural Language Processing , Cultural aware LLMs, Dialects Understanding }

\abstract{As large language models (LLMs) become increasingly central to Arabic NLP applications, their effectiveness in linguistically diverse settings, particularly regions with rich dialectal variation such as Saudi Arabia, remains underexplored. Existing evaluation paradigms tend to prioritize high-resource languages or Modern Standard Arabic (MSA), overlooking regional linguistic and cultural specificities. This leads to performance limitations and cultural biases in real-world deployments. To address this gap, we introduce \texttt{Absher}, the first comprehensive and fine-grained benchmark designed to assess the understanding of LLMs regarding Saudi dialects and their embedded cultural nuances. \texttt{Absher} consists of over 18,000 multiple choice questions derived from a curated dataset of dialectal words, phrases, and proverbs sourced from five major Saudi regions. The benchmark spans six task categories: Meaning, True/False, Fill-in-the-Blank, Contextual Usage, Cultural Interpretation, and Location Recognition, enabling multifaceted evaluation across both linguistic and cultural dimensions.We perform zero-shot evaluations on six state-of-the-art open LLMs: ALLaM, LLaMA, Jais, Mistral, Qwen, and AceGPT. Our results reveal substantial performance variability across dialects and question types. Qwen achieved the highest overall accuracy, excelling in word-level questions (63\%), while ALLaM outperformed others in the interpretation of proverbs (48\% accuracy). All models struggled with content from underrepresented dialects, particularly Southern and Eastern variants, and with context-free True/False questions, highlighting weaknesses in dialect grounding and binary reasoning. These findings demonstrate the need for dialect-aware training and culturally aligned evaluation. We position \texttt{Absher} as a critical step towards more equitable and effective LLM development for real-world Arabic applications.}

\maketitle

\section{Introduction}\label{sec:introduction}

Recent breakthroughs in the area of language processing
have shown that contemporary computational models
can perform a variety of tasks, including
text production, question answering, and machine translation \citep{Brown2020, OpenAI2023}. These powerful computational models are becoming indispensable tools across various sectors, from education and healthcare to public services and entertainment. However, a critical challenge remains: the vast majority of these models, along with their foundational training data and evaluation benchmarks, are developed with a prevailing focus on high-resource languages like English or, within the Arabic-speaking world, Modern Standard Arabic (MSA). This limits their performance and applicability in domains characterized by significant linguistic and cultural diversity, particularly those with rich regional dialects \citep{Baly2023, alwajih2025palm}. Language is inherently cultural; its full comprehension demands reasoning beyond literal meaning to grasp social implications and regional specificities.

Saudi Arabia exemplifies this challenge with its rich landscape of regional dialects. These dialects are not merely linguistic variants, but embody unique cultural practices, idiomatic expressions, and proverbs critical to authentic communication \citep{al2018suar}. Despite this profound linguistic and cultural diversity, existing Arabic natural language processing (NLP) benchmarks and large language models (LLMs) training data largely overlook these fine-grained Saudi dialectal and cultural nuances \citep{ayash2025saudiculture, Baly2023}. Therefore, models struggle to interpret local identity, leading to significant performance gaps and potentially biases against underrepresented languages.

{To address this critical gap and rigorously assess LLMs' cultural and linguistic competence in this domain, we introduce  \texttt{Absher}, a novel, comprehensive, and first-of-its-kind benchmark.  \texttt{Absher} offers an evaluation framework built on a curated dataset of authentic Saudi dialectal content, sourced from diverse regions across the Kingdom. It comprises $18,564$ multiple-choice questions, systematically generated to cover six distinct categories: Meaning, True/False, Fill-in-the-Blank, Contextual Usage, Cultural Interpretation, and Location Recognition. This multi-faceted design facilitates a deep, granular assessment of LLMs' capabilities that moves beyond superficial linguistic understanding toward genuine cultural competence}.

Beyond technical accuracy, the ability of LLMs to process and respond to Saudi dialects carries broader sociolinguistic significance. These dialects are not only linguistic systems but also carriers of regional identity, social norms, and cultural heritage. Ensuring  their equitable representation in AI systems is crucial for  linguistic inclusion and mitigating the marginalization of underrepresented communities. Such efforts are vital for the development of equitable AI systems and for informing culturally aware language technologies and public policy.

\begin{figure*}[h]
    \small
    \includegraphics[width=\textwidth, height=11cm]{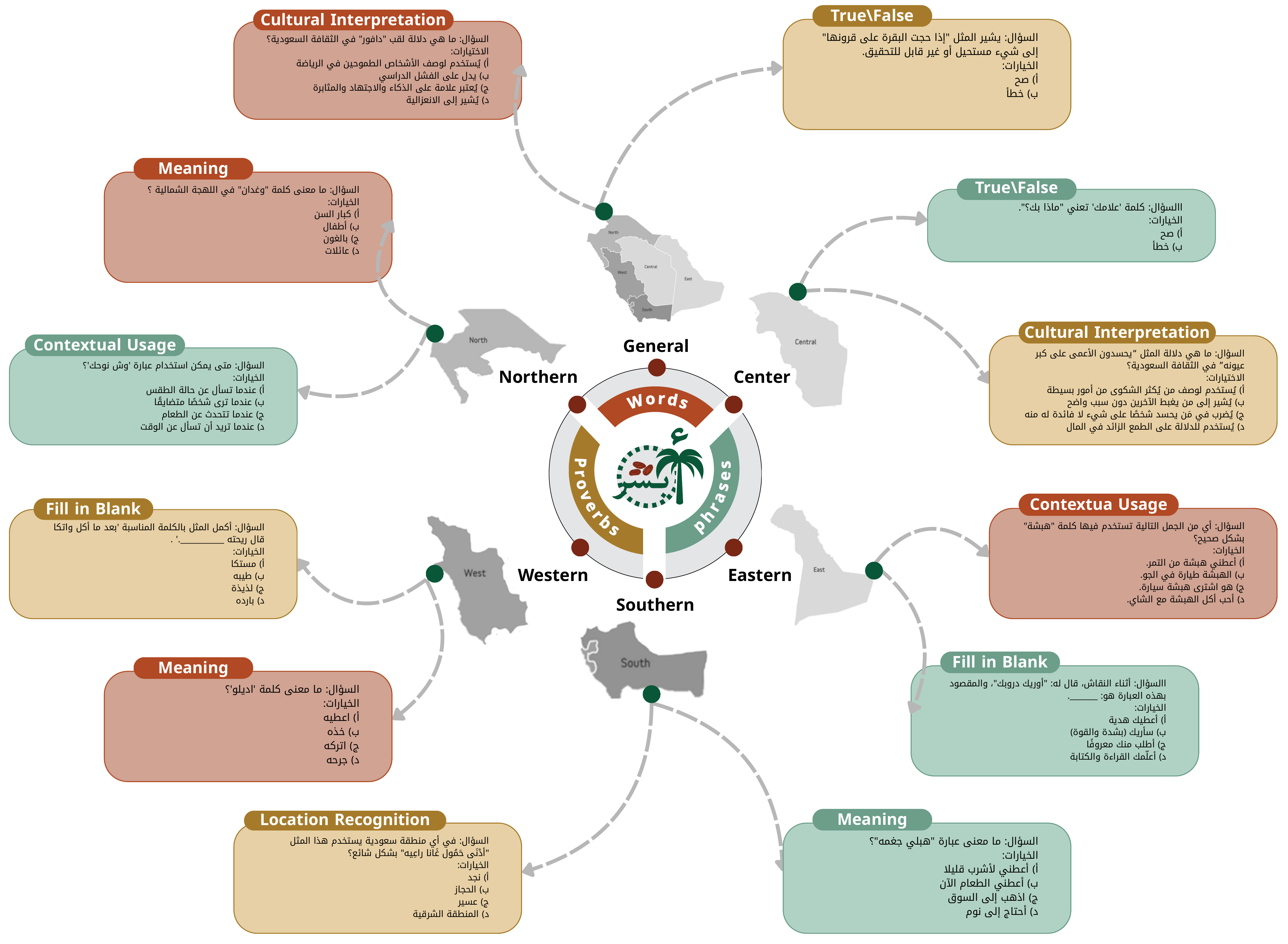} 
    \caption{Overview of \texttt{Absher},  covering all regions of Saudi Arabia (five specific regions and a general category) and presents representative examples from its three types of content: words, phrases, and proverbs. The benchmark also spans all six question types: Meaning, True/False, Fill-in-the-Blank, Contextual Usage, Cultural Interpretation, and Location Recognition.}
    \label{overview}
\end{figure*}

In this work, our objective is to provide answers to key research questions concerning LLMs' capabilities in interpreting Saudi dialects, the specific challenges they face, and a comparative analysis of their performance. Our goal is twofold: (1) to develop a novel, fine-grained benchmark that systematically evaluates LLMs across Saudi dialects and linguistic units; and (2) to conduct a comprehensive evaluation of state-of-the-art LLMs, analyzing their performance, weaknesses, and areas for improvement. Our key contributions are as follows:

\begin{itemize}
\item \textbf{Benchmark Development:} 
We introduce \texttt{Absher}, the first large-scale benchmark tailored to Saudi dialects, featuring over 18,000 multiple-choice questions. These questions span six task categories and are grounded in authentic dialectal words, phrases, and proverbs from five major Saudi regions, addressing a critical gap in Arabic NLP evaluation. Figure ~\ref{overview} illustrates the composition of \texttt{Absher} across different regions and categories. 

\item \textbf{Methodological Framework:}
We propose a scalable pipeline that combines prompt engineering with automated question generation using LLMs, followed by a rigorous multi-stage human validation process. This ensures high linguistic and cultural fidelity, making \texttt{Absher} a reliable tool for dialectal evaluation.

\item \textbf{Model Evaluation and Analysis:} 
We conduct a comprehensive evaluation of multilingual and Arabic-specific LLMs on the \texttt{Absher} benchmark. Our analysis reveals significant performance variability across dialects, task categories, and linguistic units. We also investigate the impact of model size, language orientation, and architecture, offering insights into the limitations and strengths of current LLMs in dialect-sensitive settings.

\end{itemize}

This paper is structured as follows: Section 2 reviews the existing literature on Arabic dialects in LLMs. Section 3 details the methodology for constructing the Absher benchmark, including data collection, pre-processing, prompt design, question generation, and human evaluation. Section 4 presents the experimental setup and results of our LLM evaluations. Section 5 provides a detailed analysis of our findings, including comparative performance across models, content types, and question types. Finally, Section 6 discusses the broader implications of our results and outlines promising directions for future research.

\section{Related Work}\label{relatedwork}
\subsection{Arabic Dialects in LLMs}
Recent efforts in Arabic NLP have increasingly emphasized dialectal and cultural aspects in evaluating LLMs. Several benchmarks have emerged to measure model performance in these dimensions. For example, AraDiCE \citep{Baly2023} assessed LLMs on dialectal and cultural understanding across Levantine, Egyptian, and Gulf Arabic, highlighting limitations in generation and reasoning. Calibration-Arabic-PLMs \citep{Al-Laith2025} examined model confidence in dialect identification, revealing misalignment between confidence and accuracy. Health-LLM-Arabic \citep{alharbi-etal-2025-evaluating} introduced a Cultural Sensitivity Score to evaluate model responses to health-related claims in Saudi, Egyptian, and Moroccan dialects.

In addition, to these evaluation efforts, several datasets and models have been designed to enhance dialect awareness. Dallah \citep{alwajih2024dallah} developed a multimodal assistant fine-tuned on six dialects, while AraReasoner \citep{hasanaath2025arareasoner} proposed a benchmark suite for evaluating Arabic reasoning capabilities. Jawaher \citep{magdy2025jawaher} focused on figurative language through a collection of Arabic proverbs, and CARE \citep{guo2025care} evaluated cultural alignment via human preferences in Arabic and Chinese contexts. Palm \citep{alwajih2025palm} contributed a linguistically diverse dataset covering 22 Arab countries, combining MSA and dialectal instructions.

In the multimodal space, Peacock \citep{alwajih2024peacock} released the Henna benchmark to test cultural competence, and JEEM \citep{kadaoui2025jeem} introduced dialect-aware visual tasks involving culturally grounded images. ArabicMTEB and Swan \citep{bhatia2024swan} offered resources for evaluating cross-lingual, dialectal, and domain-specific Arabic embeddings.

\subsection{Saudi Dialects in LLMs}
Despite growing interest in Arabic dialects, Saudi-specific evaluations remain limited. SaudiCulture \citep{ayash2025saudiculture} evaluated five LLMs across five Saudi regions, revealing performance drops on culturally grounded content. AL-QASIDA \citep{robinson2024qasida} examined nine LLMs across eight dialects—including Saudi Arabic—using fidelity and diglossia metrics, and discussed pretraining implications. Other benchmarks, such as Dallah \citep{alwajih2024dallah} and Palm \citep{alwajih2025palm}, included Saudi dialects but lacked focused analysis. LLM-MT-Data \citep{Elmogtaba2024} used LLMs to translate Saudi speech data into MSA, and AraBench \citep{sajjad2020arabench} explored dialect-to-English translation with fine-tuning and back-translation.

A notable data-centric effort is the KIND dataset \citep{yamani2024kind}, which collected fine-grained dialectal Arabic through a community-driven competition. KIND includes parallel corpora and Q\&A spanning 29 Saudi dialects, with city-level granularity. However, it is primarily designed for data collection and does not offer a structured evaluation of LLMs’ contextual understanding, a gap \texttt{Absher} seeks to fill.

{A key point of contrast is with SaudiCulture}~\citep{ayash2025saudiculture}{, which introduced a pioneering cultural benchmark. However, it is limited to 441 questions focused mainly on general knowledge and cultural themes predominantly in MSA. In contrast, Absher provides a broader and more targeted evaluation with a significantly larger volume of questions. It is uniquely designed for a fine-grained, dialect-sensitive assessment across Saudi dialects, focusing on the nuanced understanding of words, phrases, and proverbs. Unlike SaudiCulture, which emphasizes high-level cultural reasoning, Absher enables a deep, multi-category evaluation drawn from region-specific vernacular data, making it a more comprehensive and regionally grounded resource for advancing Arabic NLP.}

While these studies offer valuable insights, none provide a comprehensive benchmark that systematically evaluates LLMs on Saudi dialects across linguistic units such as words, phrases, and proverbs. Existing work often overlooks the sociocultural specificity of regional expressions. In contrast, \texttt{Absher} introduces a fine-grained, culturally contextualized benchmark that spans multiple Saudi dialects and domains, enabling detailed evaluation of LLMs' dialectal and cultural understanding. It further differentiates itself through real-world question design, multi-unit coverage, and region-aware analysis of model behavior.

\section{Constructing Absher Benchmark}

\begin{figure}[h]
    \centering
    \includegraphics[width=1.0\textwidth]{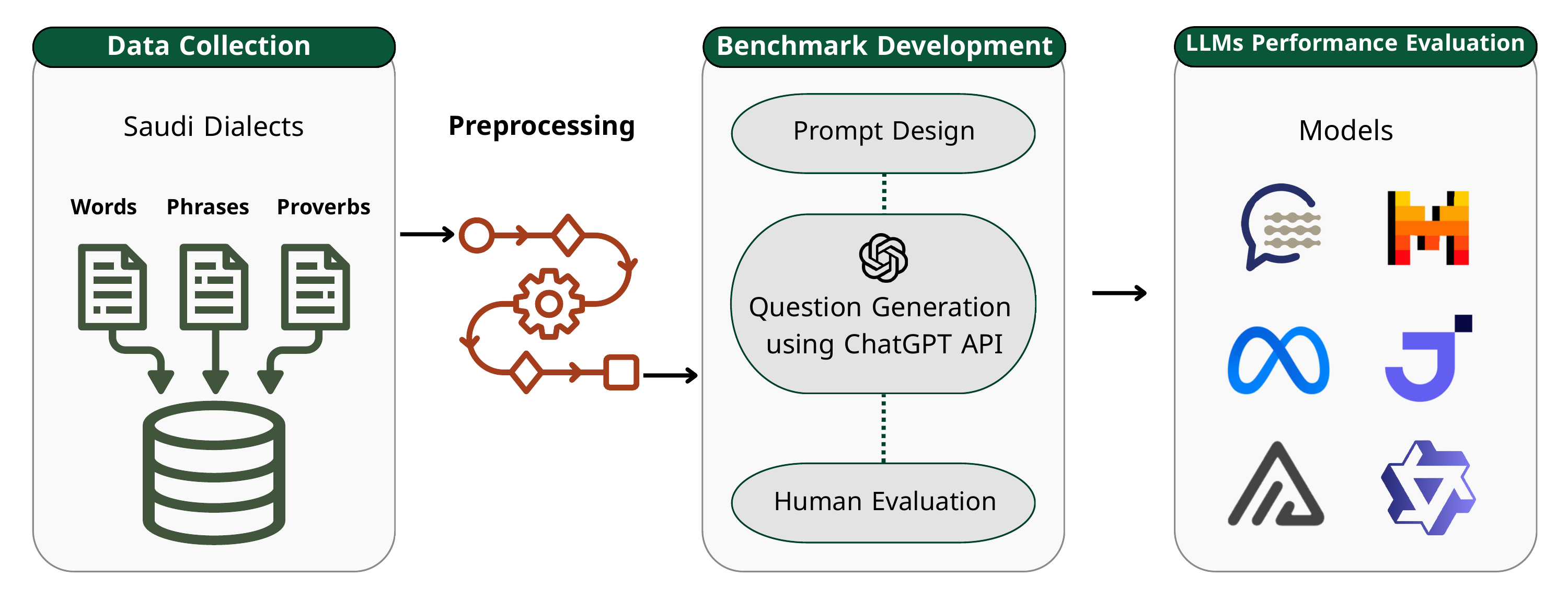} 
    \caption{{The overall pipeline of constructing \texttt{Absher} benchmark.}}
    \label{fig:pipeline}
\end{figure}

To assess the ability of computational systems to understand the dialectal and cultural nuances specific to Saudi Arabia, we introduce the \texttt{Absher} Benchmark. This evaluation framework is built upon a curated set of dialectal contents, including words, phrases, and proverbs, representing diverse regions across Saudi Arabia. {The development process followed a structured pipeline encompassing data collection, pre-processing, prompt engineering, automated question generation, and expert-based review as seen in Figure}~\ref{fig:pipeline}.

The resulting benchmark provides a systematic and scalable method for evaluating LLMs on their ability to interpret regional language variation, contextual meanings, and culturally embedded expressions. As illustrated in Figure~\ref{fig:pipeline}, the development process spans from initial data acquisition to final model evaluation, allowing both fine-grained analysis and performance benchmarking of LLMs in Saudi dialects.

\subsection{Data Collection}

To evaluate the ability of computational models to interpret Saudi dialects and culturally embedded expressions, we introduce \texttt{Absher}, a benchmark grounded in authentic regional content. It is built from a curated collection of dialectal items, including words, phrases, and proverbs, representing linguistic diversity across Saudi Arabia. 

The core of the dataset was collected from the Moajam website\footnote{\url{https://ar.mo3jam.com/}}, a digital repository that aggregates dialectal vocabulary across Arabic regions. We scraped and organized entries into three categories: words, phrases, and proverbs, each annotated with its meaning, dialect label, contextual usage, and translation. This process ensured a comprehensive coverage of both literal and figurative expressions relevant to regional Saudi Arabic.

\begin{itemize}
    \item \textbf{Words} refer to vocabulary items specific to Saudi dialects, often including slang or regional variants not found in Modern Standard Arabic. 

    \item \textbf{Phrases} encompass idiomatic or conversational expressions commonly used across regions. 

    \item \textbf{Proverbs} are culturally rooted sayings that convey implicit meanings, requiring contextual and cultural understanding. 
    
\end{itemize}

The raw collection initially consisted of 5,483 items, which were refined through pre-processing and filtering, resulting in 3,094 unique contents used to generate over 18,000 multiple-choice questions across six task types. Table~\ref{tab:generated-questions} summarizes the dataset statistics.

\begin{table*}[ht]
\caption{{Content statistics before and after pre-processing and the number of generated questions by category.}}
\centering
\begin{tabular}{lccc}
\toprule
\textbf{Category} & \textbf{\# Contents (Raw)} & \textbf{\# Contents (Filtered)} & \textbf{\# Generated Qs} \\
\midrule
Words     & 4,428  & 2,533 & 15,198 \\
Phrases   & 869    & 478   & 2,868  \\
Proverbs  & 186    & 83    & 498    \\
\midrule
\textbf{Total} & 5,483 & 3,094 & \textbf{18,564} \\
\bottomrule
\end{tabular}
\label{tab:generated-questions}
\end{table*}

\subsection{Pre-processing Steps}

After data collection, we applied a systematic pre-processing pipeline to refine the dataset for effective question generation. This phase included deduplication, normalization, and structural formatting. We focused exclusively on entries labeled as Saudi dialects, retaining regional variants such as Central, Western, Southern, Northern, and Eastern. Additionally, general Saudi terms—expressions widely used across the country regardless of region—were preserved to ensure comprehensive linguistic coverage. We further standardized spelling and usage variations across dialects to ensure consistency, particularly for words with multiple colloquial forms. This harmonization was essential for generating uniform prompts and minimizing ambiguity in model evaluation. After filtering and normalization, the dataset was reduced to 3,094 contents: 2,533 words, 478 phrases, and 83 proverbs, as shown in Table~\ref{tab:generated-questions}. Finally, the cleaned data was structured into a machine-readable format suitable for automated prompt generation, forming a high-quality foundation for the benchmark tasks.

\subsection{Prompt Design}

To rigorously assess the capabilities of LLMs in interpreting Saudi dialects, we employed a structured prompt design strategy tailored to dialectal understanding. Each prompt template included four core components: a task description, a dialect-based question, multiple-choice answer options, and a designated correct answer. Table~\ref{tample-ex} provides an illustration. The prompts were derived from dialectal contents, such as words, phrases, and proverbs. Each annotated with its dialect label, meaning, and a contextual usage example.

The initial prompt generation followed a simple formulation process, but several issues emerged. Correct answers frequently appeared in fixed positions (e.g., always option A), True/False questions were heavily biased toward ``True'', and some questions included incorrect dialectal labels. These inconsistencies introduced unintended biases, undermining the fairness and reliability of the benchmark.

To address these limitations, we refined the prompt design by introducing structural controls. The correct answer positions were randomized to prevent positional bias, and the distribution of True/False items was balanced to ensure fairness. Dialect annotations were reviewed and validated by linguistic experts to confirm their accuracy.  {Crucially, we implemented a rigorous two-step filtering strategy to handle abnormal model outputs from GPT-4o:}
 
\begin{itemize}
\item \textbf{Automated Filtering (Formatting and Structure):} {All generated outputs were subjected to a validation script to check for strict structural integrity. This script automatically flagged and discarded any questions that contained malformed JSON output, failed to include exactly four distinct multiple-choice options, or did not designate a clear correct answer. This ensured only well-structured, testable items proceeded to the next phase.}

\item \textbf{Manual Curation (Hallucination and Semantic Errors):} {A team of two native Saudi Arabic speakers and linguistic experts manually reviewed a sample of automatically validated questions. Their primary task was to filter out semantic errors and hallucinations, specifically removing questions where the generated options were culturally inaccurate, the question’s premise contradicted the source material (meaning or context), or the designated correct answer was objectively wrong.}
\end{itemize}

These refinements significantly improved the robustness, validity, and representativeness of the final set of multiple-choice questions, establishing \texttt{Absher} as a dependable tool for dialect-sensitive LLMs evaluation. For the complete prompt template, please refer to Appendix~\ref{sec:prompt}. 

\begin{table*}[ht]
\caption{{Template for question formulation}}
\centering
\begin{tabular}{ll}  
    \toprule
    \textbf{Component} & \textbf{Description} \\
    \midrule
    \textbf{Task:} & Describes the task or the function of the question. \\
    \textbf{Question:} & The question that is presented to the model.  \\
    \textbf{Multi-choice Options:} &  Option A, B, C, and D.  \\ 
    \textbf{True/False Options:} 
    & A) True, B) False  \\
    \textbf{Correct Answer:} & The correct answer to the question. \\
    \bottomrule
\end{tabular}
\label{tample-ex}
\end{table*}

\subsection{Question Generation}
Building upon the refined prompt templates, we conducted large-scale question generation using the \textbf{ChatGPT API (GPT-4o)}. We selected GPT-4o for this task due to its strong multilingual capabilities and demonstrated proficiency in both literal and contextual understanding. Its robustness in handling diverse linguistic inputs made it particularly well-suited for generating high-quality, dialect-specific multiple-choice questions that require nuanced cultural reasoning. Compared to smaller or monolingual models, GPT-4o also exhibited greater consistency and fewer hallucinations—qualities essential for ensuring the validity of benchmark items.

Each cleaned and pre-processed content item, whether a word, phrase, or proverb, was programmatically paired with its dialect label, meaning, and usage example to construct a complete input prompt. This setup enabled GPT-4o to generate questions that evaluated not only surface-level understanding but also deeper comprehension of dialectal and cultural nuances. For each dialectal item, the model generated six distinct types of questions, with each type targeting a specific dimension of language understanding:

\begin{itemize}
\item \textbf{Meaning:} Assessing the model’s ability to correctly interpret or define dialectal expressions.
\item \textbf{True/False:} Testing the factual correctness of statements about the term or its usage.
\item \textbf{Fill-in-the-Blank:} Evaluating the model’s ability to infer missing words in context.
\item \textbf{Contextual Usage:} Determining whether the model can accurately apply the term within a complete sentence.
\item \textbf{Cultural Interpretation:} Probing the understanding of cultural or metaphorical meanings, particularly in proverbs.
\item \textbf{Location Recognition:} Identifying the geographic or regional origin associated with the dialectal term.
\end{itemize}

This systematic generation approach ensured comprehensive coverage of the linguistic and cultural characteristics embedded in the Saudi dialects. By applying the process across all content categories, words, phrases, and proverbs, a total of \textbf{18,564} multiple-choice questions were generated. These questions form the core of the \textit{Absher} benchmark and allow for a rigorous, multifaceted evaluation of LLM performance. Table~\ref{tab:questions-per-dialect} shows the distribution of questions across regional dialects, highlighting the benchmark’s broad linguistic representation.

{The distribution of questions across regional dialects reflects the linguistic landscape of Saudi Arabia. The Najdi dialect, spoken predominantly in the Central region, and the Hejazi dialect, prevalent in the Western region, are spoken by a larger portion of the population and receive extensive exposure through media and online platforms. This widespread usage provides a richer pool of linguistic material. By contrast, the Southern, Northern, and Eastern dialects have a more localized presence, with fewer readily accessible resources. Despite these differences, the benchmark includes all major dialects, thus capturing the full spectrum of Saudi Arabic and supporting a comprehensive evaluation of LLM performance.}

\begin{table}[h]
\caption{{Number of questions per regional dialects in the \textit{Absher} benchmark.}}
\centering
\small
\setlength{\tabcolsep}{6pt} 
\begin{tabular}{lr}
\toprule
\textbf{Dialect} & \textbf{\# Qs} \\
\midrule
{Central} & 5,652 \\
Western & 5,262 \\
Southern & 2,682 \\
Northern & 414 \\
Eastern & 216 \\
General Saudi Terms & 4,338 \\
\bottomrule
\end{tabular}
\label{tab:questions-per-dialect}
\end{table}

\begin{figure*}[h]
    \small
    \includegraphics[width=13cm, trim=4.5cm 5cm 4cm 2.1cm, clip]{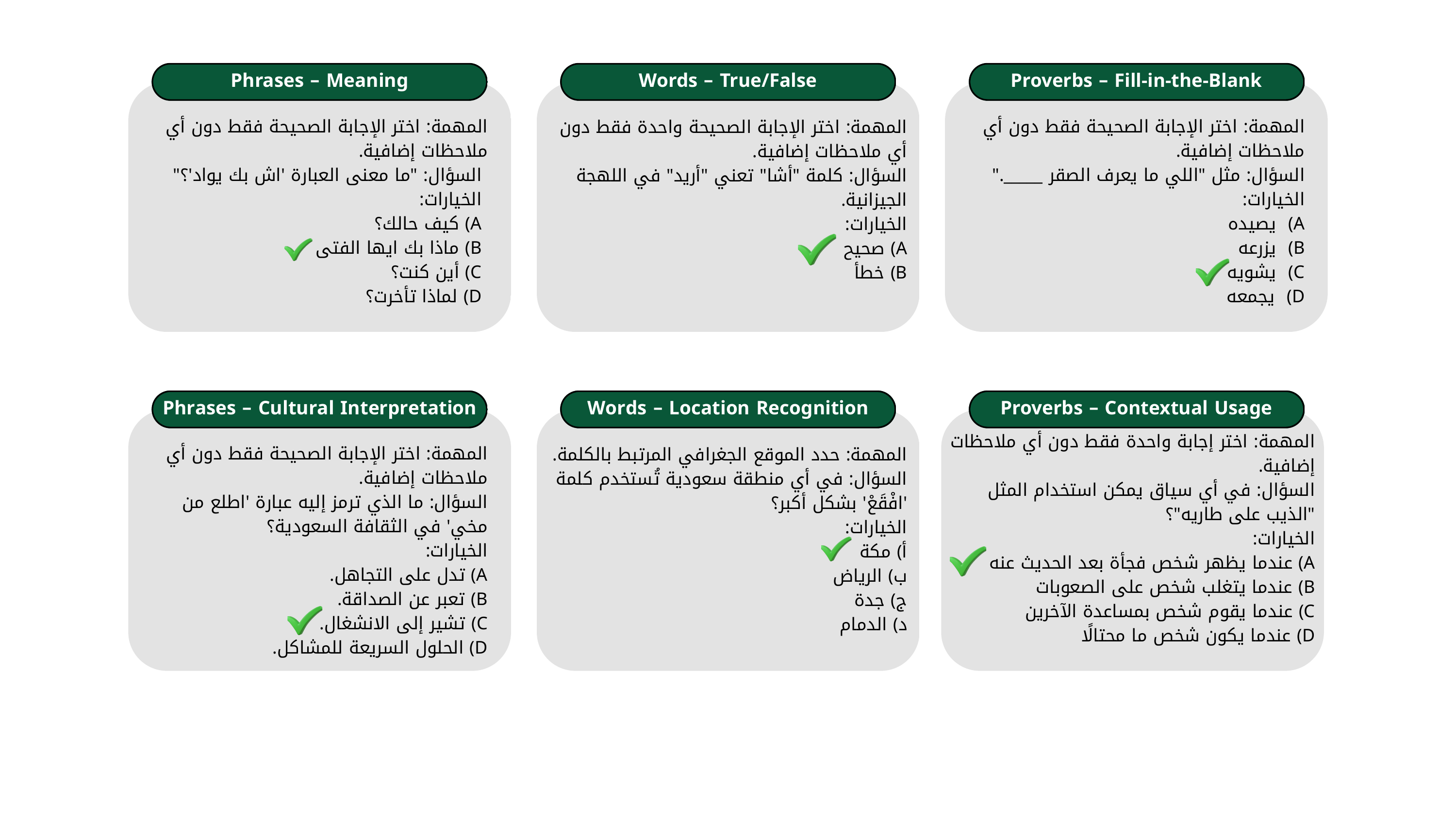} 
    \caption{An illustration of the different question types. The first line in each example indicates the task type, the second line presents the question, and the subsequent lines list the answer options. The correct answers are marked in green.}
    \label{examples6}
\end{figure*}

Figure~\ref{examples6} illustrates the six types of questions included in the benchmark. Together, these types enable a well-rounded evaluation of model capabilities, covering meaning interpretation, cultural and regional understanding, and recognition of local expressions. The use of a standardized prompt template throughout the generation process ensures consistency, minimizes bias, and strengthens the reliability of the evaluation. 


\subsection{Human Evaluation}

To validate the reliability, correctness, and cultural appropriateness of the questions in the \texttt{Absher} benchmark, we performed a structured human evaluation. This step was essential to ensure that the automatically generated content adhered to linguistic standards and dialectal relevance, especially given the complexity and regional variability of Saudi dialects. Four native Saudi Arabic speakers participated in the evaluation. All annotators were fluent in multiple regional dialects and possessed strong familiarity with the cultural and linguistic diversity across Saudi Arabia. They were divided into two independent groups, each assigned a distinct subset of the data. This setup enabled broader dataset coverage, minimized cross-group bias, and reduced cognitive fatigue.

A pilot study preceded the main annotation phase to ensure consistent interpretation of the evaluation guidelines. Annotators were presented with 20 representative sample contents (120 questions total), covering diverse dialects and question types. These examples served to familiarize them with the annotation tasks and resolve any ambiguities in advance. To test attentiveness and accuracy, the pilot set included quality control questions with plausible yet incorrect answers. Annotators, who misclassified these, were received follow-up clarifications to align their understanding with the evaluation standards. This step improved consistency and reliability of the judgment in the formal evaluation.

The main evaluation involved a stratified random sample comprising 10\% of the benchmark contents, i.e., Words, Phrases, and Proverbs, across all six question types. This resulted in 600 annotated questions. Complete annotation guidelines, including task definitions, assessment criteria, and illustrative examples, are provided in Appendix~\ref{sec:annotation_guideline}. Based on these instructions, annotators completed two core tasks:

\begin{itemize}

\item \textbf{Question Quality:} Evaluate each question for clarity, fluency, and logical structure, with attention to dialectal appropriateness and cultural sensitivity.

\item \textbf{Answer Correctness:} Verify whether the answer labeled as correct by the model aligns with the ground-truth answer based on the dialectal dataset.
\end{itemize}

Inter-annotator agreement was measured using Cohen’s Kappa coefficient. Table~\ref{tab:kappa_scores} shows strong agreement across most question types, particularly for Meaning and Cultural Interpretation, where Kappa values exceed 0.94. While Fill-in-the-Blank and Contextual Usage tasks yielded slightly lower agreement in question quality (0.71 and 0.72, respectively), the values still fall within acceptable reliability thresholds. Additionally, we computed the overall alignment between model-generated answers and human judgments, which reached 93.03\%, confirming the reliability and consistency of the benchmark’s human evaluation process.

\begin{table}[h]
\centering
\small 
\caption{{Cohen's Kappa scores for question quality and answer correctness across different question types. \textbf{Bold} denotes the top scores.}}
\label{tab:kappa_scores}
\begin{tabular}{lcc}
\toprule
\textbf{Q-Type} & \textbf{Q-Quality} & \textbf{A-Correctness} \\
\midrule
Meaning & \textbf{0.97} & \textbf{0.97} \\
True/False & 0.95 & 0.87 \\
Fill-in-the-Blank & 0.71 & 0.91 \\
Contextual Usage & 0.72 & 0.88 \\
Cultural Interpretation & 0.83 & 0.94 \\
Location Recognition & 0.85 & 0.92 \\
\bottomrule
\end{tabular}
\end{table}

\section{Experiments}


We evaluated a set of open-source LLMs: \textbf{LLaMA-3 8B Instruct} \citep{Touvron2024}, \textbf{Jais-13B} \citep{sengupta2023}, \textbf{ALLaM-7B} \citep{bari2025allam}, \textbf{Mistral-7B} \citep{jiang2023}, \textbf{Qwen2.5-7B Instruct} \citep{yang2024qwen2p5}, and \textbf{AceGPT-7B-chat} \citep{huang2023acegpt}. These models were chosen based on three main criteria: open availability, Arabic language support, and competitive performance across a range of NLP tasks. All models were evaluated in a {zero-shot setting}, without any task-specific fine-tuning, to assess their inherent ability to understand Saudi dialects and cultural content. Since the benchmark is composed exclusively of multiple-choice questions, we employed four standard classification metrics: {Accuracy}, {Precision}, {Recall}, and {F1 Score}. These metrics offer a comprehensive and balanced evaluation of model performance, in line with prior work on zero-shot LLMs assessment for low-resource languages \citep{lowres_llm_eval_2025}.

\begin{table*}[h]
\caption{Average evaluation results of models performance. {\textbf{Bold}} denotes the best result achieved across all models for each content type, while {\underline{underlining}} indicates the best-performing content type for each individual model.}
\centering
\begin{tabular}{llcccc}
\toprule
\textbf{Model} & \textbf{Content Type} & \textbf{Accuracy} & \textbf{Precision} & \textbf{Recall} & \textbf{F1-score}  \\
\midrule
\multirow{4}{*}{\includegraphics[width=0.7cm]{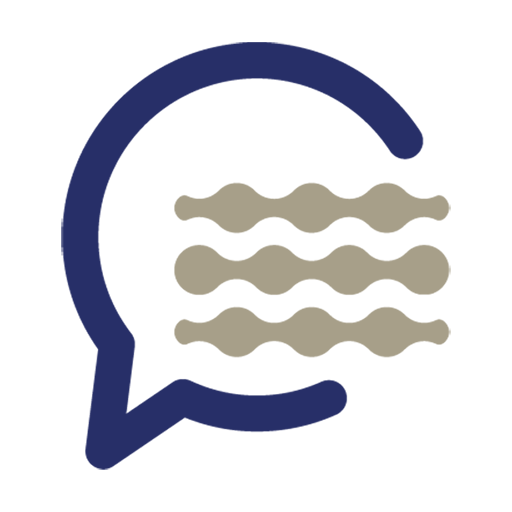} \textbf{ALLaM}} 
& Word & 40.32\% & 33.07\% & 26.63\% & 17.28\% \\
& Phrase & 42.28\% & 30.54\% & 26.38\% & 19.42\% \\
& Proverb & \besttype{\bestmodel{48.74\%}} & 39.95\% & 37.76\% & 31.43\% \\
\midrule
\multirow{4}{*}{\includegraphics[width=0.8cm]{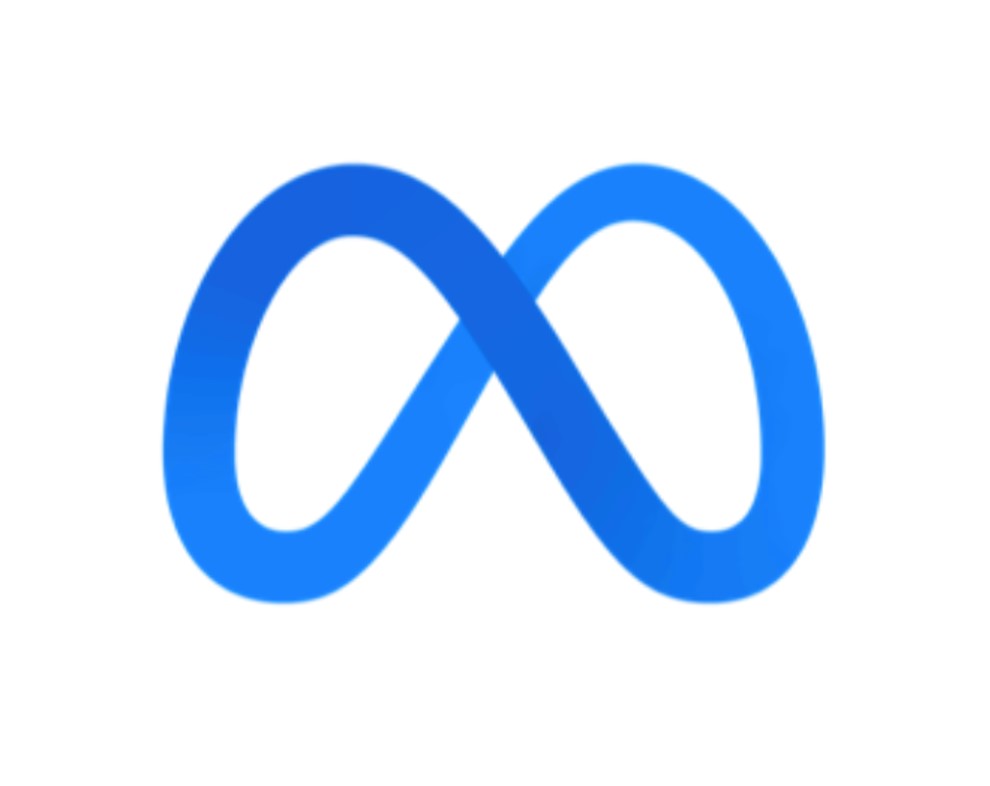} \textbf{LLaMA}} 
& Word & 41.30\% & 36.19\% & 23.99\% & 16.63\% \\
& Phrase & 43.41\% & 29.66\% & 24.13\% & 18.07\% \\
& Proverb & \bestmodel{45.31\%} & 28.55\% & 28.60\% & 18.87\% \\
\midrule
\multirow{4}{*}{\includegraphics[width=0.8cm]{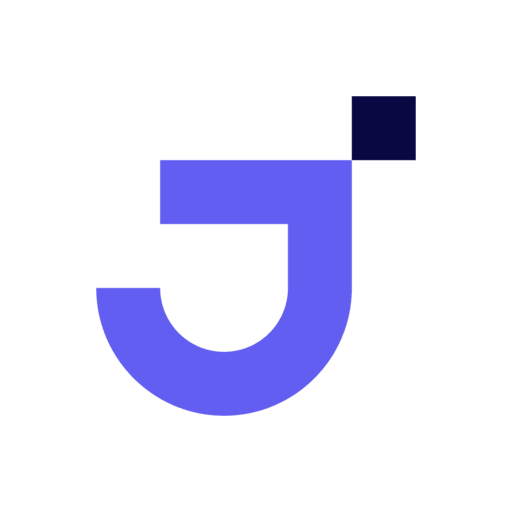} \textbf{Jais}} 
& Word & \bestmodel{42.66\%} & 32.78\% & 24.76\% & 14.78\% \\
& Phrase & 41.94\% & 37.81\% & 27.27\% & 16.27\% \\
& Proverb & 41.81\% & 17.52\% & 28.33\% & 15.91\% \\
\midrule
\multirow{4}{*}{\includegraphics[width=0.5cm]{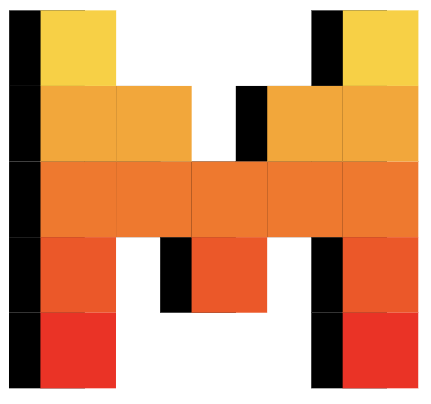} \textbf{Mistral}} 
& Word & \bestmodel{47.90\%} & 22.80\% & 30.05\% & 23.98\% \\
& Phrase & \besttype{44.80\%} & 28.11\% & 33.09\% & 26.02\% \\
& Proverb & 45.10\% & 18.27\% & 29.05\% & 22.31\% \\
\midrule
\multirow{4}{*}{\includegraphics[width=0.7cm]{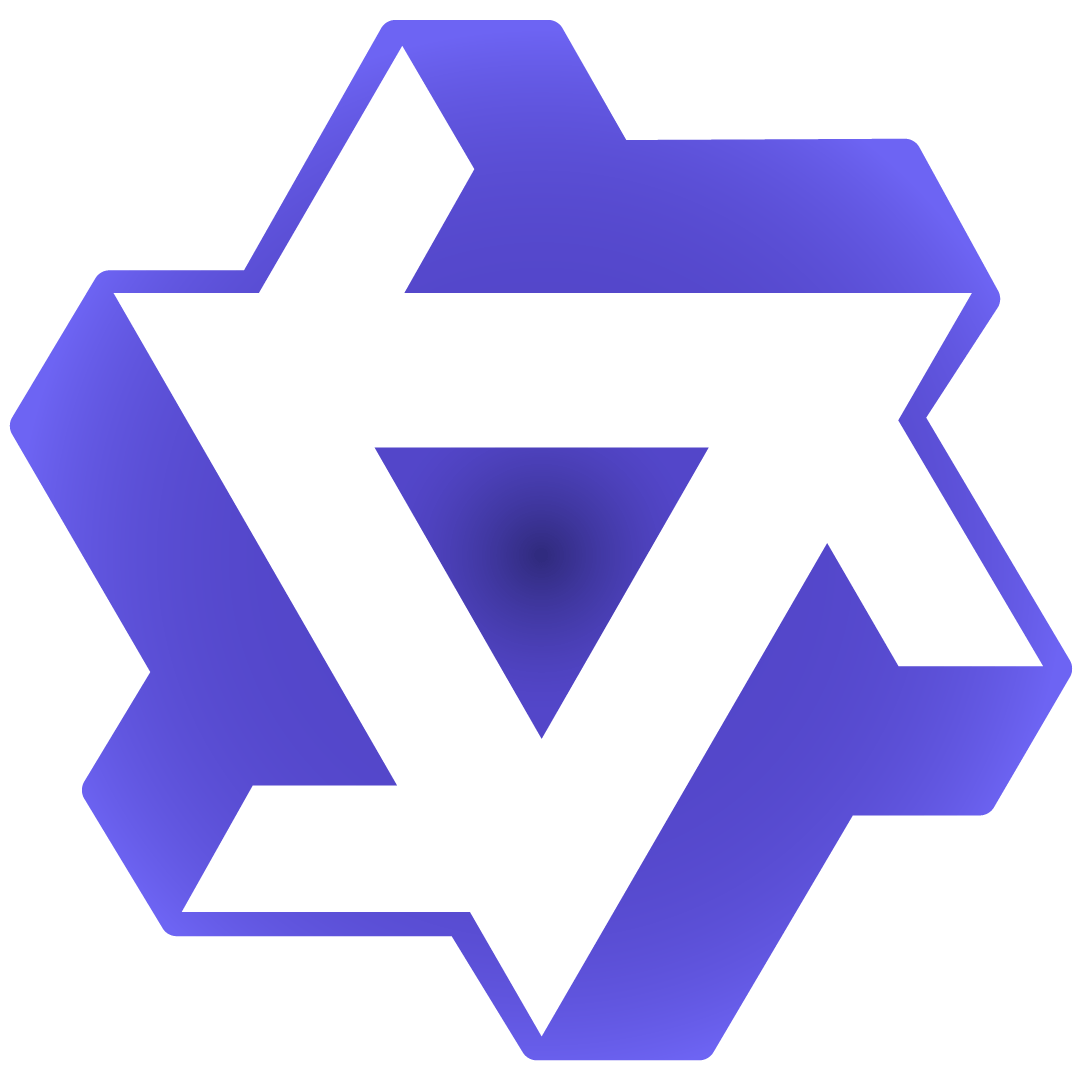} \textbf{Qwen}} 
& Word & \besttype{\bestmodel{63.00\%}} & 42.07\% & 46.10\% & 41.18\% \\
& Phrase & 43.68\% & 33.58\% & 32.93\% & 26.00\% \\
& Proverb & 44.38\% & 31.36\% & 33.16\% & 29.56\% \\
\midrule
\multirow{4}{*}{\includegraphics[width=0.7cm]{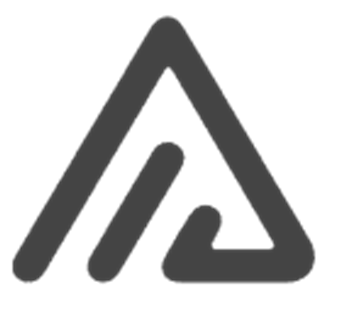} \textbf{ACEGPT}} 
& Word & \bestmodel{41.90\%} & 33.98\% & 33.18\% & 28.30\% \\
& Phrase & 39.93\% & 33.68\% & 32.55\% & 29.61\% \\
& Proverb & 34.81\% & 37.93\% & 33.15\% & 27.51\% \\
\bottomrule
\end{tabular}%
\centering
\label{tab:metrics_average_total}
\end{table*}

\subsection{Results}

Table~\ref{tab:metrics_average_total} presents the performance of the six evaluated models, such as ALLaM, LLaMA, Jais, Mistral, Qwen, and ACEGPT, across different content types, namely words, phrases, and proverbs. The results reveal noticeable variations in performance depending on both the linguistic content and the complexity of the task\footnote{The full set of results is provided in Appendix~\ref{sec:Detailed-Performance}.}. For word-level content, the Qwen model demonstrated the strongest capabilities, achieving an accuracy of 63.00\% and an F1-score of 41.18\%. These results indicate Qwen's effectiveness in handling isolated lexical items, suggesting its advantage in processing tasks that rely on clear, context-independent meanings. 

{To contextualize the reported performance}, it is important to note the chance-level baseline for this benchmark. Since most multiple-choice questions contain four options, random guessing corresponds to 25\% accuracy (and 50\% for True/False items). Therefore, the observed performance range of approximately 40–60\% accuracy across models is substantially above chance, highlighting both the non-trivial capability of the evaluated LLMs and the inherent difficulty of the dialect- and culture-sensitive questions in Absher.

In contrast, phrase-level tasks presented a more competitive landscape. Mistral achieved the highest F1-score at 26.02\%, with Qwen and ACEGPT following closely behind. This reflects a broader challenge for all models when dealing with phrases, which often carry more ambiguous or context-dependent meanings compared to individual words. Proverb-based tasks emerged as the most challenging across all models, reflecting the complexity and cultural depth embedded in such expressions. Despite these challenges, ALLaM led in this category with an accuracy of 48.74\% and an F1-score of 31.43\%, indicating its relative strength in handling idiomatic and culturally grounded language. In contrast, models such as Jais consistently recorded lower F1-scores across all content types, particularly for proverbs, highlighting limitations in their ability to process culturally rich or metaphorical expressions.

These findings show a consistent trend: while certain models demonstrate strong performance on words or phrases, accurately processing complex cultural constructs like proverbs remains a significant challenge. This highlights the need for further research to enhance LLMs' ability to handle the nuanced and context-heavy aspects of language, particularly those rooted in cultural knowledge and idiomatic usage.

\section{Analysis}


\subsection{Impact of Model Size and Language Orientation}

Table~\ref{tab:model_combined} summarizes the models evaluated with respect to parameter size, language orientation, and open-source availability. The set of models includes Arabic-specific and multilingual LLMs, spanning different sizes and scales, from 7B to 13B parameters.

\begin{table*}[h]
\caption{Overview of the evaluated models, highlighting their size, supported languages, and average accuracy achieved in the experiments. The best results are listed from highest to lowest, with the three best results in bold.}
\small 
\centering
\begin{tabular}{lccc}
\toprule
\textbf{Model} & \textbf{Size} & \textbf{Language} & \textbf{Accuracy} \\
\midrule
\multirow{0.5}{*}{\includegraphics[width=0.3cm]{Qwen.png} \textbf{Qwen2.5 Instruct }} 
& 7B & Multilingual & \textbf{50.35\%} \\
\midrule
\multirow{0.5}{*}{\includegraphics[width=0.3cm]{M-logo.png} \textbf{Mistral}} 
& 7B & Multilingual & \textbf{45.93\%} \\
\midrule
\multirow{0.5}{*}{\includegraphics[width=0.3cm]{ALLaM1.png} \textbf{ALLaM {3rd}}} 
& 7B & Arabic & \textbf{43.78\%} \\
\midrule
\multirow{0.5}{*}{\includegraphics[width=0.4cm]{lama.jpg} \textbf{LLaMA-3 Instruct}} 
& 8B & Multilingual & 43.34\% \\
\midrule
\multirow{0.5}{*}{\includegraphics[width=0.3cm]{jais.png} \textbf{Jais}} 
& 13B & Arabic & 42.14\% \\
\midrule
\multirow{0.5}{*}{\includegraphics[width=0.3cm]{ACE.png} \textbf{AceGPT-chat}} 
& 7B & Arabic & 38.88\% \\
\bottomrule
\end{tabular}
\label{tab:model_combined}
\end{table*}



While it is often assumed that larger parameter counts lead to better performance, our results demonstrate that accuracy does not scale linearly with size. For example, Qwen, a 7B multilingual model, achieved the highest overall accuracy of 50.35\%, outperforming the larger Arabic-native Jais model (13B) by nearly 10\%. Similarly, Mistral (7B) achieved 45.93\%, exceeding the performance of some larger Arabic-specialized models. A deeper comparison of Arabic-native models (ALLaM, Jais, ACEGPT) versus multilingual LLMs (Qwen, Mistral, LLaMA) reveals additional insights. Despite being trained specifically on Arabic data, Arabic-native models did not consistently outperform their multilingual counterparts in tasks involving dialectal variation and cultural nuances.

Qwen demonstrated strong performance across tasks, particularly in contextual usage (64.7\%) and cultural interpretation (63.7\%), highlighting the benefits of multilingual pretraining combined with advanced architecture. ALLaM, though smaller and Arabic-focused, performed competitively in proverb-based cultural interpretation (60.0\%) and contextual usage tasks. Mistral, a multilingual model with only 7B parameters, achieved balanced performance and notably excelled in binary reasoning (58.4\% in word-level True/False questions), emphasizing the importance of architectural optimization. LLaMA showed clear strengths in location recognition (66.3\%) but underperformed in tasks requiring deeper semantic understanding. Among Arabic-native models, ACEGPT achieved solid results in contextual usage (54.5\%), while Jais led in location recognition (67.6\%) but exhibited limitations in cultural and semantic tasks.

Overall, these findings suggest that specialization in Arabic does not inherently guarantee superior performance on culturally complex or dialectally rich tasks. Instead, the combination of diverse multilingual training, effective architectural design, and optimized parameter utilization appears to be more influential. Performance tends to be stronger on lexical and recognition tasks but decreases as linguistic and cultural complexity increases, particularly with proverbs and culturally loaded expressions.

\subsection{Examination of Model Responses by Region}

\begin{figure}[h]
    \centering
    \includegraphics[width=\textwidth]{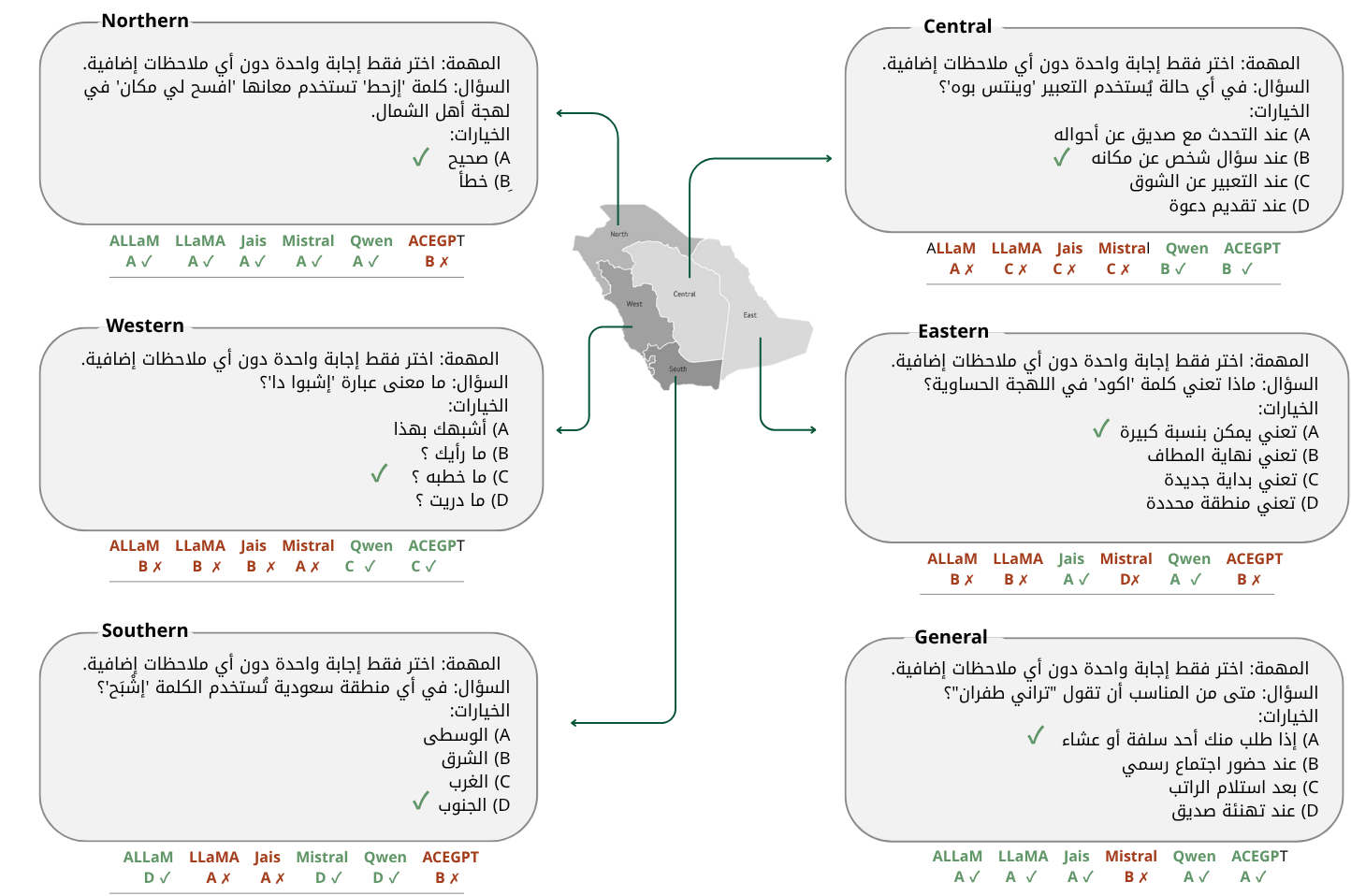}
    \caption{Model responses to dialect-based questions from different Saudi regions. Each example shows the question, its regional, and the answers from six LLMs.}
    \label{fig:dialect_responses}
\end{figure}

Our evaluation of LLMs responses to region-specific linguistic questions, detailed in Figure~\ref{fig:dialect_responses}, reveals considerable performance variability influenced by both the geographic origin of expressions and the nature of the linguistic task. Notably, we observed strong performance in the {General} category for a widely known proverb, with five out of six models providing correct answers. This aligns with the expectation that frequently cited expressions, which appear widely in public discourse, are more likely to be accurately interpreted due to their higher representation in pretraining datasets.

In contrast, expressions from the {Southern} region resulted in a consistent failure across all six models. The proverb tested was regionally specific and less documented, suggesting limited representation in training data and highlighting the persistent challenges LLMs face with culturally embedded content from low-resource dialects. For the {Central} region, performance on a location recognition task was mixed. While ALLaM and Jais correctly identified the dialectal origin, other models defaulted to major urban centers such as Riyadh or Jeddah. This pattern likely reflects an urban-centric bias within the models' training corpora. Interestingly, the {Northern} region, despite being underrepresented in online resources, saw three models correctly answer a true/false phrase question. This suggests that task structure can, to some extent, mitigate data scarcity, as binary questions allow models to leverage contextual inference or statistical patterns to produce correct responses.

In the {Eastern} region, performance remained low, with only ACEGPT correctly interpreting the contextual usage of a time-related word. This outcome emphasizes the difficulty models encounter when handling dialectal expressions that rely on contextual understanding, particularly for less represented regions. Finally, for the {Western} region, five of six models successfully identified the cultural function of a widely circulated annoyance phrase. This likely reflects the increased presence of such expressions in social or digital spaces, enhancing their visibility within pre-training data.

Overall, these results reveal persistent challenges in capturing region-specific linguistic and cultural nuances. Although models tend to perform well with broadly known or globally circulated expressions, they consistently struggle with dialects and phrases deeply rooted in local practices. This underscores the need for more inclusive and dialect-rich datasets that reflect the full diversity of linguistic and cultural realities between regions.

\subsection{Performance Across Content and Question Types}

\begin{figure}[h]
    \centering
    \includegraphics[width=\textwidth]{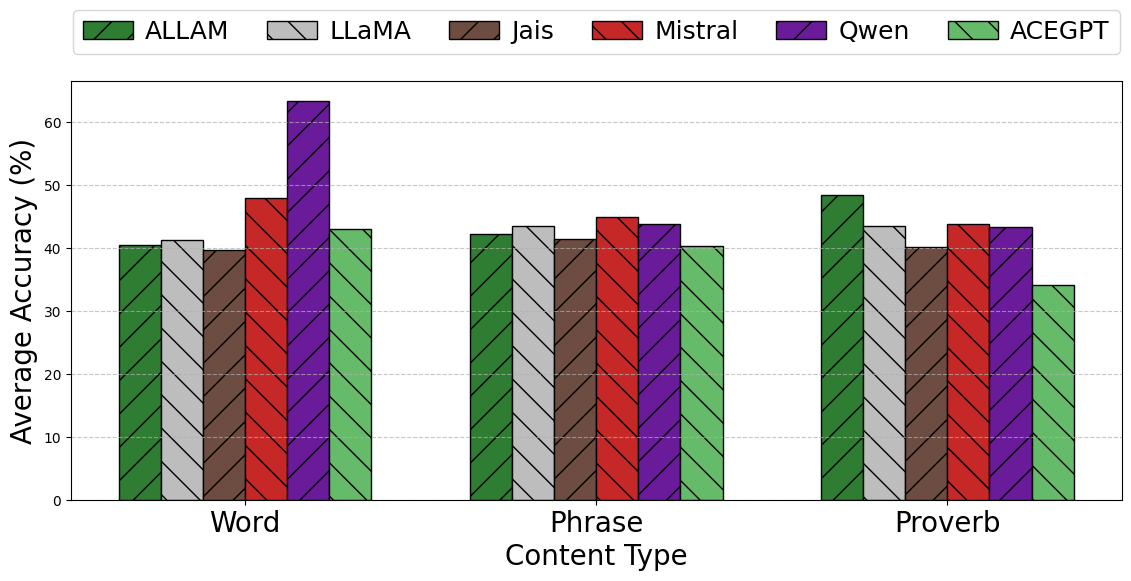} 
    \caption{Average Accuracy for Different Models by Content Type}
    \label{fig:Accuracy_Models}
\end{figure}

Figure~\ref{fig:Accuracy_Models} presents a detailed analysis of model performance across content categories, revealing key trends in how different LLMs handle dialectal material. Qwen consistently outperformed other models, achieving the highest accuracy on word-level questions, particularly in the meaning and contextual usage types, with an overall peak of approximately 63.5\%. ALLaM demonstrated notable strengths in cultural interpretation tasks, especially when dealing with proverbs, where it surpassed 48\% accuracy. An interesting pattern emerged in relation to content type: questions based on proverbs consistently resulted in higher model accuracy compared to those based on isolated words or phrases. This may be attributed to the richer context and more explicit cultural cues embedded within proverbs, which provide clearer semantic and pragmatic signals.These results suggest that LLMs benefit from contextually grounded input and perform more reliably when exposed to culturally meaningful expressions.

\begin{figure}[h]
    \centering
    \includegraphics[width=\textwidth]{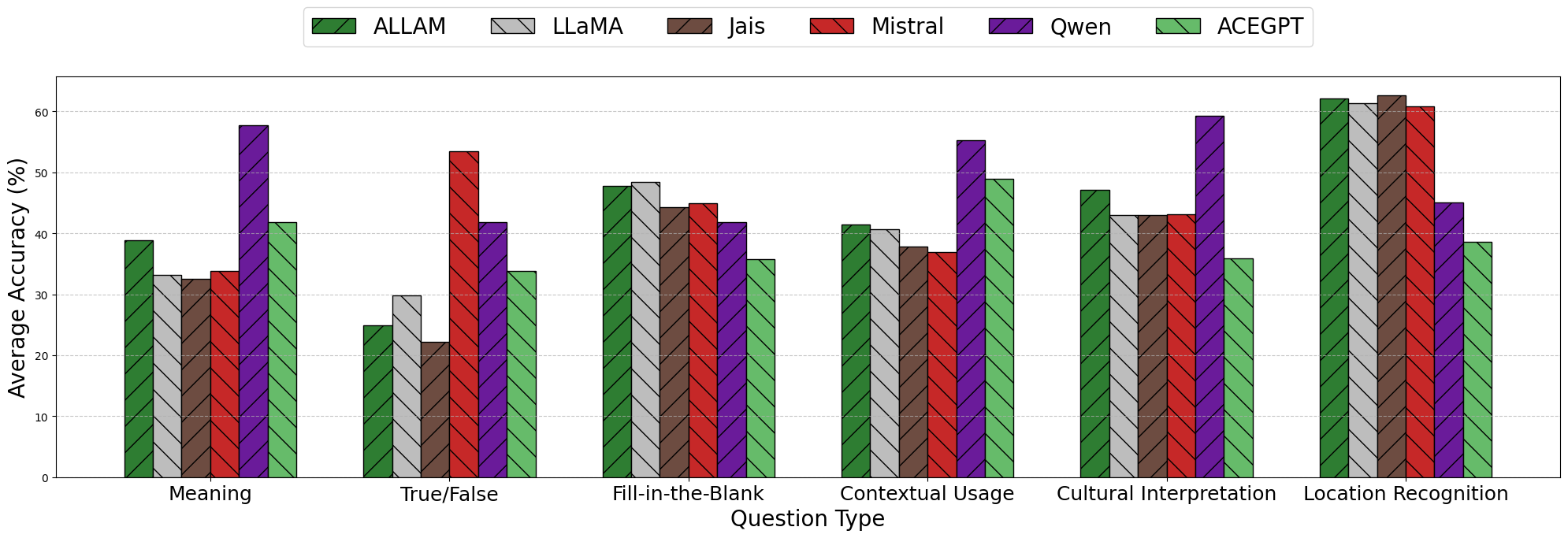} 
    \caption{Average Accuracy for Different Models by Question Type}
    \label{fig:model_performance2}
\end{figure}

Complementing this content-based analysis, Figure~\ref{fig:model_performance2} illustrates how the models performed on different types of questions. Among all types, location recognition questions were the most successfully handled, with average model accuracy exceeding 58\%. Jais performed particularly well in this category, reaching 62\% accuracy—indicating that models are generally capable of linking dialectal features to their corresponding regions. In contrast, the True/False questions pose the greatest challenge, especially when derived from short or context-poor input, such as individual words or brief phrases. These questions required binary reasoning with limited context, leading to lower performance. Accuracy for this type ranged from 22\% (Jais) to 58\% (Qwen), underscoring the difficulties LLMs face in making confident decisions under ambiguity or minimal input cues. Overall, these results highlight the dual importance of linguistic structure and contextual richness in benchmarking the dialectal and cultural understanding of LLMs.


\subsection{Case Study of LLMs Responses Across Question Types and Dialects}
Our case study investigates the performance of six LLMs across culturally embedded tasks derived from various Saudi dialectal regions. To illustrate observed performance patterns, we examine specific model responses across varied content types, question structures, and dialects, as detailed in Tables \ref{tab:example-proverb-baqara} to \ref{tab:example-question}. The results indicate that regionally common and widely used expressions are more accurately interpreted by LLMs. For example, in the general category, the proverb: "Itha Hajat Al-Baqara Ala Qoroniha", its translation is "If the cow performs Hajj on its horns" (See Table \ref{tab:example-proverb-baqara}) was correctly understood by five out of six models: ALLaM, LLaMA, Jais, Qwen, and ACEGPT all provided the correct answer, while only Mistral failed. This proverb is relatively widespread, and its meaning, ``something impossible to achieve'', is often documented in public sources, likely increasing its representation in model training data. In contrast, proverbs that are more regionally isolated or culturally specific, such as the Southern fill-in-the-blank proverb : "Khattet fi maa wa aqbas fi heid", its translation is "Plan in water and pinch in a rock" (See Table \ref{tab:example-proverb-blank}), saw a complete failure across all six models. None of the models, i.e., ALLaM, LLaMA, Jais, Mistral, Qwen, and ACEGPT, selected the correct option, suggesting a consistent gap in exposure to underrepresented regional idioms, especially those originating from areas with fewer accessible textual resources.


\begin{table*}[h]
\caption{{Model answers to a proverb meaning question in the general dialect.}}
\centering
\scalebox{0.85}{ 
\begin{tabularx}{\textwidth}{>{\bfseries}l X}
\toprule
Content Type:      & Proverb \\
Question Type:     & Meaning \\
Region:            & General \\
\midrule
\multicolumn{2}{c}{
\includegraphics[width=\linewidth]{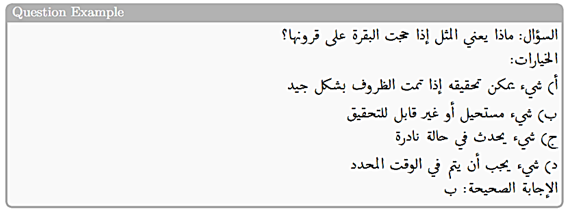}
} \\
\midrule
Models' Answer: &
\correct{ALLaM1.png}{ALLaM: B}, 
\correct{lama.jpg}{LLaMA: B}, 
\correct{jais.png}{Jais: B}, 
\incorrect{M-logo.png}{Mistral: C}, 
\correct{Qwen.png}{Qwen: B}, 
\correct{ACE.png}{ACEGPT: B} \\
\bottomrule
\end{tabularx}}
\label{tab:example-proverb-baqara}
\end{table*}



\begin{table*}[h]
\caption{{Model answers to a fill-in-the-blank proverb in the southern dialect.}}
\centering
\scalebox{0.85}{ 
\begin{tabularx}{\textwidth}{>{\bfseries}l X}
\toprule
Content Type:      & Proverb \\
Question Type:     & Fill-in-the-Blank \\
Region:            & Southern \\
\midrule
\multicolumn{2}{c}{
\includegraphics[width=1.0\linewidth]{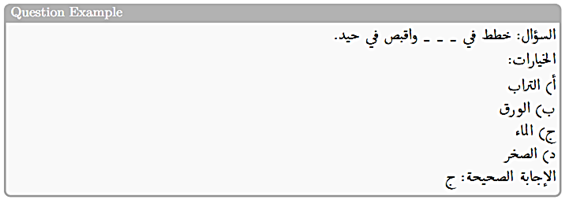}
} \\
\midrule
Models' Answer: &
\incorrect{ALLaM1.png}{ALLaM: D}, 
\incorrect{lama.jpg}{LLaMA: B}, 
\incorrect{jais.png}{Jais: D}, 
\incorrect{M-logo.png}{Mistral: D}, 
\incorrect{Qwen.png}{Qwen: A}, 
\incorrect{ACE.png}{ACEGPT: D} \\
\bottomrule
\end{tabularx}}
\label{tab:example-proverb-blank}
\end{table*}


\begin{table*}[h]
\caption{{Model answers to a location recognition word question in the central dialect.}}
\centering 
\scalebox{0.85}{
\begin{tabularx}{\textwidth}{>{\bfseries}l X}
\toprule
Content Type:      & Word \\
Question Type:     & Location Recognition \\
Region:            & Central \\
\multicolumn{2}{c}{
\includegraphics[width=1.0\linewidth]{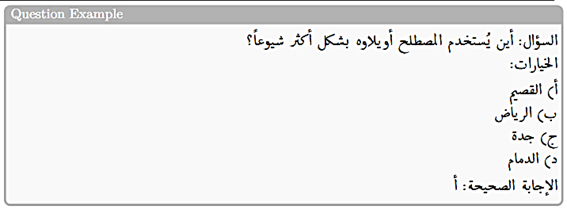}
} \\
\midrule
Models' Answer: &
\correct{ALLaM1.png}{ALLaM : A},
\incorrect{lama.jpg}{LLaMA : B}, 
\correct{jais.png}{Jais : A}, 
\incorrect{M-logo.png}{Mistral : B}, 
\incorrect{Qwen.png}{Qwen : C}, 
\incorrect{ACE.png}{ACEGPT : D} \\
\bottomrule
\end{tabularx}}
\label{tab:example-question}
\end{table*}

Additionally, the study revealed notable differences in the models’ ability to comprehend regional dialects, particularly in how they responded to specific tasks based on region and question type. For example, when tasked with identifying a location-specific term (See Table \ref{tab:example-question}) from the Central region, only two models, ALLaM and Jais, were able to correctly associate it with the Al-Qassim region. The other models, however, defaulted to more familiar urban areas, like Riyadh and Jeddah, likely due to the prevalence of these dialects in the training materials. Urban dialects from cities such as Riyadh, Jeddah, and Khobar are more commonly represented in online resources, which could explain why the models showed a preference for these regions. In contrast,  like Qassim, Aseer, or the southern regions face challenges due to limited textual data from these areas.

Moreover, when it comes to other local dialects from the Eastern, Northern, or Southern parts of Saudi Arabia, significant discrepancies in the models’ performance were noted. Expressions and words specific to these regions were often misunderstood, highlighting the gap in representation for these dialects. This contrast underscores the fact that while widely recognized expressions are easily interpreted, culturally unique idioms that are specific to certain regions are much harder for the models to grasp.

Overall, our comprehensive evaluation underscores that while contemporary LLMs possess considerable linguistic breadth in Arabic dialects, their depth of understanding for more nuanced, regionally isolated, or culturally embedded expressions remains a significant limitation. These findings collectively point to the crucial need for more extensive and geographically diverse dialectal data in future LLMs training to achieve truly robust and culturally sensitive linguistic AI.

\section{Discussion}
This research presents \texttt{Absher}, a novel benchmark aimed at evaluating LLMs on Saudi dialects, a linguistically rich yet underrepresented subset of Arabic. Unlike prior work such as Dallah \cite{alwajih2024dallah}, Palm \cite{alwajih2025palm}, and AL-QASIDA \cite{robinson2024qasida}, which aggregate dialects or focus on broader Arabic settings, \texttt{Absher} uniquely provides fine-grained, region-specific coverage of Saudi dialects, including Central, {Western}, Southern, Eastern, and Northern varieties. Absher's strength lies in its foundation: over 18,000 multiple-choice questions derived from a curated dataset of dialectal words, phrases, and proverbs sourced from these five major Saudi regions. The benchmark systematically spans six distinct task categories: Meaning, True/False, Fill-in-the-Blank, Contextual Usage, Cultural Interpretation, and Location Recognition.

Our findings highlight notable patterns in how LLMs handle Saudi dialects. Location Recognition tasks yielded the highest performance, likely due to the concrete and frequently occurring nature of geographic references in training data, which aids retrieval and identification. In contrast, True/False questions were the most challenging, especially in low-context settings, reflecting the inherent difficulty of binary reasoning tasks that require nuanced cultural and contextual understanding beyond surface-level lexical cues.

Furthermore, Qwen’s strong performance, especially in Meaning, Contextual Usage, and Cultural Interpretation questions, demonstrates its robust semantic understanding and ability to accurately interpret dialectal expressions in context. {Across content types}, performance varied by model rather than being uniformly highest for any single type. While some models (e.g., ALLaM) achieved relatively stronger results on Proverbs, others, such as Qwen, performed best on Word-level questions, indicating that certain models benefit more from the richer contextual and cultural cues provided by longer and more nuanced inputs. For instance, ALLAM’s relatively high performance in Cultural Interpretation of proverbs highlights its ability for deeper socio-cultural reasoning, likely attributed to its Arabic-centric training.

The results reveal a noticeable gap between multilingual and Arabic-native models. Models like Qwen and Mistral, which were trained on a wide range of languages, outperformed their Arabic-centric counterparts in overall accuracy. Their ability to generalize across dialectal variations, despite not being trained explicitly on Saudi dialects, reflects the benefits of exposure to diverse language data. These results align with observations from Palm \cite{alwajih2025palm} and related studies, where multilingual models correlated with higher robustness on non-standard Arabic input.

Nevertheless, Arabic-native models demonstrated strengths in specific areas. For example, ALLaM achieved high accuracy in tasks involving proverbs, while Jais performed relatively well in location-based recognition. These patterns suggest that domain-specific pretraining provides advantages in interpreting traditional or idiomatic expressions, even if it does not generalize as effectively across diverse question types. Such findings align with results from the AraDiCE benchmark \cite{Baly2023}, where models exhibited improved performance when test data resembled the training distribution. To further demonstrate model differences, consider a Cultural Interpretation question based on the proverb: "Ally ma ya'arf alsger yshwaeh", meaning "He who doesn't recognize the falcon cooks it" 
While ALLAM accurately selected the metaphorical interpretation (“Those who lack knowledge may misuse valuable things”), other models like Jais and ACEGPT selected overly literal or generic options. This reflects ALLAM's stronger grounding in socio-cultural expressions—a critical strength when applying LLMs in culturally sensitive settings.

In summary, our findings reveal a clear gap in current LLMs’ ability to comprehensively handle the diverse spectrum of Saudi Arabic dialects, especially in less standardized or culturally nuanced forms. While multilingual models demonstrate stronger generalization capabilities, Arabic-native models show targeted strengths in specific areas such as cultural interpretation. However, neither category achieves consistently strong performance across all dialects and task types. {This discrepancy can be attributed to structural imbalances in the representation of dialectal forms within the training data, where widely documented or urban-centered dialects are more prevalent, whereas regionally specific or culturally embedded expressions remain underrepresented. Furthermore, performance patterns indicate that model success is influenced not only by linguistic coverage but also by the alignment between task type and the type of cultural knowledge required. These insights underscore the need for more representative and balanced datasets, alongside tailored adaptation strategies and controlled evaluation methods, to improve model alignment with dialectal Arabic at the national level and to better understand why certain models excel in some contexts while struggling in others.}

\subsection{Limitations}
While the \texttt{Absher} benchmark represents a significant step toward evaluating LLMs on Saudi dialects and culture, the study is subject to several limitations. First, the evaluations were conducted under constrained computational resources, which limited our ability to explore multiple model configurations or conduct repeated trials. Second, all models were assessed using a fixed prompt structure (e.g., “Answer as if you are a Saudi…”), which may have influenced model behavior. Future work could explore the impact of prompt variation and instruction tuning. Third, although human evaluation was performed on a stratified sample to ensure dataset quality, a complete manual review of all 18,564 questions was not feasible due to the scale and diversity of the benchmark. Finally, {while Absher covers various regional Saudi dialects, the representation of these dialects within the dataset is not perfectly balanced; for instance, Central and Western dialects are more prevalent. That highlights the need for more diverse and evenly distributed resources to comprehensively capture all Saudi dialectal variations in future research.} Future research should leverage more extensive computational infrastructure, experiment with prompt engineering strategies and fine-tuning methods. It should also incorporate broader expert reviews to further enhance the accuracy, reliability, and cultural validity of the benchmark evaluations.

\subsection{Ethical Considerations}

This study utilizes publicly available data sources to evaluate the dialectal understanding of LLMs within the Saudi context. Care was taken to ensure that no private, identifiable, or sensitive information was included in the dataset. {To guarantee this, all dialectal content—including words, phrases, and proverbs— was sourced exclusively from  the publicly accessible Moajam website. This resource functions as a digital repository of aggregated dialectal vocabulary and cultural expressions. We explicitly excluded any conversational data, social media content, or personally generated user text to safeguard against the unintentional inclusion of private information or the leakage of toxic language that is often prevalent in online datasets.}

{The primary ethical risk in dialectal evaluation is the potential for linguistic discrimination, where underrepresented dialects are implicitly or explicitly associated with lower social status or performance failure by the model. To mitigate this critical risk, we implemented two concrete measures. First, the Absher dataset was rigorously designed to ensure fair coverage across the five major Saudi regions (Central, Western, Southern, Eastern, and Northern), deliberately challenging the common bias in NLP datasets that over-represent urban or dominant dialects. Second, the evaluation tasks focus strictly on testing linguistic competence (meaning, contextual usage, and cultural interpretation) rather than social or demographic inference, preventing the model from being scored on tasks that could encourage or propagate negative stereotypes associated with any specific regional dialect.}


Given that language models may produce hallucinated or inaccurate outputs—particularly in low-resource or culturally nuanced scenarios—we stress that these outputs should not be used for critical decision-making without human oversight. The evaluation framework was carefully designed to minimize cultural bias and to avoid reinforcing stereotypes. 


{Finally, our commitment to responsible AI usage extended to our human validation processes. The validation, which was performed by native speakers from each of the five regions to confirm dialectal authenticity, was governed. Throughout the project, we maintained a commitment to transparency and responsible AI usage, acknowledging the inherent limitations of current LLMs, with the ultimate goal of promoting the development of fair, inclusive, and culturally aware language technologies.}

\subsection{Conclusion and Future Work}



This paper introduced \texttt{Absher}, a novel benchmark designed to evaluate the capabilities of LLMs in understanding the diverse dialects and rich cultural expressions across Saudi Arabia. Unlike existing Arabic benchmarks that focus on Modern Standard Arabic (MSA) or treat dialects at a coarse level, \texttt{Absher} offers fine-grained, region-specific evaluation through over 18,000 carefully constructed multiple-choice questions. These questions span six task types—ranging from meaning identification to cultural interpretation—and are grounded in authentic dialectal content from various Saudi regions. These questions are derived from a curated dataset of dialectal words, phrases, and proverbs sourced from five major Saudi regions, and systematically span six distinct task categories: Meaning, True/False, Fill-in-the-Blank, Contextual Usage, Cultural Interpretation, and Location Recognition. This comprehensive design enables multifaceted evaluation across both linguistic and cultural dimensions.

Our evaluation of several state-of-the-art LLMs revealed significant variation in performance across dialects, question types, and linguistic units. While some models performed well on tasks like location recognition and contextual usage, they struggled with cultural inference and binary reasoning, especially in low-context scenarios. Notably, multilingual models such as Qwen and Mistral outperformed Arabic-native models in general tasks, whereas the latter showed an advantage in interpreting culturally embedded expressions like proverbs.

Beyond technical performance, this work contributes to a broader sociolinguistic goal: fostering inclusive and culturally sensitive NLP systems. By systematically incorporating dialectal diversity and cultural context, \texttt{Absher} helps bridge existing evaluation gaps and supports the development of language technologies that reflect the lived linguistic realities of users.


{Looking forward, future work will expand \texttt{Absher} to include additional dialects, question formats, and model types. We plan to conduct more in-depth evaluations using alternative strategies, such as few-shot and fine-tuning scenarios, to better explore models’ capabilities across dialectal variations and cultural contexts. In addition, we aim to perform deeper causal analyses to understand the factors driving model performance across dialects and task types, providing clearer insights into how linguistic and cultural features influence LLM behavior. Integrating sociolinguistic features and region-specific datasets may further improve alignment with real-world language use. Through these efforts, we aim to promote more inclusive, robust, and context-aware Arabic NLP systems.}

\section*{Data Availability}
We provided several illustrative examples from our benchmark in this paper, and the complete dataset will be made publicly available for research purposes upon acceptance.

\vspace{0.5cm}

\noindent \textbf{CRediT authorship contribution statement}

\textbf{Renad Al-Monef}: Methodology, Data Curation, Software, Validation, Formal analysis, Visualization, Writing - original draft. \textbf{Hassan Alhuzali}: Conceptualization, Supervision, Validation, Writing - review \& editing.  \textbf{Nora Alturayeif}: Supervision, Writing - review \& editing. \textbf{Ashwag Alasmari}: Conceptualization, Supervision, Validation, Writing - review \& editing.

\section*{Declaration of competing interest}
The authors declare that they have no competing interests.


\section*{Acknowledgment}

Declaration of generative AI and AI-assisted technologies in the writing process: During the preparation of this work, we used GPT-4, an AI chatbot developed by OpenAI, to improve our written work. After using this tool/service, we reviewed and edited the content as needed and take full responsibility for the content of the publication.


The authors would like also to extend their appreciation to the Deanship of Research and Graduate Studies at \textbf{King Khalid University} for funding this work through small group research under grant number \textbf{RGP1/69/46}. The authors also express their sincere gratitude to Al Bandari Ali Alqahtani , Zainab Abdulaziz Dalim for their invaluable support and insightful contributions during the evaluation of Absher benchmark


\bibliography{sn-bibliography} 

@inproceedings{yamani2024kind,
  title={The kind dataset: A social collaboration approach for nuanced dialect data collection},
  author={Asma Yamani and Raghad Alziyady and Reem AlYami and Salma Albelali and Leina Albelali and Jawharah Almulhim and Amjad Alsulami and Motaz Alfarraj and Rabeah Al-Zaidy},
  booktitle={Proceedings of the 18th conference of the European chapter of the association for computational linguistics: Student research workshop},
  pages={32--43},
  year={2024}
}

@article{Brown2020,
author = {Brown, Tom B. and Mann, Benjamin and Ryder, Nick and Subbiah, Melanie and Kaplan, Jared and Dhariwal, Prafulla and Neelakantan, Arvind and Shyam, Pranav and Sastry, Girish and Askell, Amanda and Agarwal, Sandhini and Herbert-Voss, Ariel and Krueger, Gretchen and Henighan, Tom and Child, Rewon and Ramesh, Aditya and Ziegler, Daniel M. and Wu, Jeffrey and Winter, Clemens and Hesse, Christopher and Chen, Mark and Sigler, Eric and Litwin, Mateusz and Gray, Scott and Chess, Benjamin and Clark, Jack and Berner, Christopher and McCandlish, Sam and Radford, Alec and Sutskever, Ilya and Amodei, Dario},
title = {Language models are few-shot learners},
year = {2020},
isbn = {9781713829546},
publisher = {Curran Associates Inc.},
address = {Red Hook, NY, USA},
booktitle = {Proceedings of the 34th International Conference on Neural Information Processing Systems},
articleno = {159},
numpages = {25},
location = {Vancouver, BC, Canada},
series = {NIPS '20}
}

@misc{OpenAI2023,
  author    = {OpenAI},
  title     = {GPT-4 Technical Report},
  year      = {2023},
  url       = {https://openai.com/research/gpt-4}
}

@article{Baly2023,
title = "{A}ra{D}i{CE}: Benchmarks for Dialectal and Cultural Capabilities in {LLM}s",
    author = "Mousi, Basel  and
      Durrani, Nadir  and
      Ahmad, Fatema  and
      Hasan, Md. Arid  and
      Hasanain, Maram  and
      Kabbani, Tameem  and
      Dalvi, Fahim  and
      Chowdhury, Shammur Absar  and
      Alam, Firoj",
    editor = "Rambow, Owen  and
      Wanner, Leo  and
      Apidianaki, Marianna  and
      Al-Khalifa, Hend  and
      Eugenio, Barbara Di  and
      Schockaert, Steven",
    booktitle = "Proceedings of the 31st International Conference on Computational Linguistics",
    month = jan,
    year = "2025",
    address = "Abu Dhabi, UAE",
    publisher = "Association for Computational Linguistics",
    url = "https://aclanthology.org/2025.coling-main.283/",
    pages = "4186--4218"
}

@article{al2018suar,
title = {SUAR: Towards Building a Corpus for the Saudi Dialect},
journal = {Procedia Computer Science},
volume = {142},
pages = {72-82},
year = {2018},
note = {Arabic Computational Linguistics},
issn = {1877-0509},
doi = {https://doi.org/10.1016/j.procs.2018.10.462},
url = {https://www.sciencedirect.com/science/article/pii/S187705091832163X},
author = {Nora Al-Twairesh and Rawan Al-Matham and Nora Madi and Nada Almugren and Al-Hanouf Al-Aljmi and Shahad Alshalan and Raghad Alshalan and Nafla Alrumayyan and Shams Al-Manea and Sumayah Bawazeer and Nourah Al-Mutlaq and Nada Almanea and Waad Bin Huwaymil and Dalal Alqusair and Reem Alotaibi and Suha Al-Senaydi and Abeer Alfutamani}
}

@misc{Sengupta2023,
  author = {Sengupta, Neha and Sahu, Sunil Kumar and Jia, Bokang and Katipomu, Satheesh and Li, Haonan and Koto, Fajri and Afzal, Osama Mohammed and Kamboj, Samta and Pandit, Onkar and Pal, Rahul and Pradhan, Lalit and Mujahid, Zain Muhammad and Baali, Massa and Fikri Aji, Alham and Liu, Zhengzhong and Hock, Andy and Feldman, Andrew and Lee, Jonathan and Jackson, Andrew and Nakov, Preslav and Baldwin, Timothy and Xing, Eric},
  title = {Jais-13b},
  year = {2023},
  publisher = {Hugging Face},
  howpublished = {\url{https://huggingface.co/inceptionai/jais-13b}},
  url = {https://huggingface.co/inceptionai/jais-13b},
  note = {Accessed: 2025-08-1}
}

@article{lowres_llm_eval_2025,
title = {Evaluation of open and closed-source LLMs for low-resource language with zero-shot, few-shot, and chain-of-thought prompting},
journal = {Natural Language Processing Journal},
volume = {10},
pages = {100124},
year = {2025},
issn = {2949-7191},
doi = {https://doi.org/10.1016/j.nlp.2024.100124},
url = {https://www.sciencedirect.com/science/article/pii/S2949719124000724},
author = {Zabir Al Nazi and Md. Rajib Hossain and Faisal Al Mamun}
}

@misc{Touvron2024,
  author    = {Touvron, H. and Lavril, T. and Izacard, G. and others},
  title     = {LLaMA 3: Open Foundation and Instruction-Tuned Models},
  year      = {2024},
  note      = {Meta AI.}
}

@misc{yang2024qwen2p5,
    title = {Qwen2.5: A Party of Foundation Models},
    url = {https://qwenlm.github.io/blog/qwen2.5/},
    author = {An Yang and Baosong Yang and Beichen Zhang and Binyuan Hui and Bo Zheng and Bowen Yu and Chengyuan Li and Dayiheng Liu and Fei Huang and Haoran Wei and Huan Lin and Jian Yang and Jianhong Tu and Jianwei Zhang and Jianxin Yang and Jiaxi Yang and Jingren Zhou and Junyang Lin and Kai Dang and Keming Lu and Keqin Bao and Kexin Yang and Le Yu and Mei Li and Mingfeng Xue and Pei Zhang and Qin Zhu and Rui Men and Runji Lin and Tianhao Li and Tianyi Tang and Tingyu Xia and Xingzhang Ren and Xuancheng Ren and Yang Fan and Yang Su and Yichang Zhang and Yu Wan and Yuqiong Liu and Zeyu Cui and Zhenru Zhang and Zihan Qiu and others},
    month = {September},
    year = {2024}
}

@article{huang2023acegpt,
    title = "{A}ce{GPT}, Localizing Large Language Models in {A}rabic",
    author = "Huang, Huang  and
      Yu, Fei  and
      Zhu, Jianqing  and
      Sun, Xuening  and
      Cheng, Hao  and
      Dingjie, Song  and
      Chen, Zhihong  and
      Alharthi, Mosen  and
      An, Bang  and
      He, Juncai  and
      Liu, Ziche  and
      Chen, Junying  and
      Li, Jianquan  and
      Wang, Benyou  and
      Zhang, Lian  and
      Sun, Ruoyu  and
      Wan, Xiang  and
      Li, Haizhou  and
      Xu, Jinchao",
    editor = "Duh, Kevin  and
      Gomez, Helena  and
      Bethard, Steven",
    booktitle = "Proceedings of the 2024 Conference of the North American Chapter of the Association for Computational Linguistics: Human Language Technologies (Volume 1: Long Papers)",
    month = jun,
    year = "2024",
    address = "Mexico City, Mexico",
    publisher = "Association for Computational Linguistics",
    url = "https://aclanthology.org/2024.naacl-long.450/",
    doi = "10.18653/v1/2024.naacl-long.450",
    pages = "8139--8163"
}

@inproceedings{bari2025allam,
    title={{ALL}aM: Large Language Models for Arabic and English},
    author={M Saiful Bari and Yazeed Alnumay and Norah A. Alzahrani and Nouf M. Alotaibi and Hisham Abdullah Alyahya and Sultan AlRashed and Faisal Abdulrahman Mirza and Shaykhah Z. Alsubaie and Hassan A. Alahmed and Ghadah Alabduljabbar and Raghad Alkhathran and Yousef Almushayqih and Raneem Alnajim and Salman Alsubaihi and Maryam Al Mansour and Saad Amin Hassan and Dr. Majed Alrubaian and Ali Alammari and Zaki Alawami and Abdulmohsen Al-Thubaity and Ahmed Abdelali and Jeril Kuriakose and Abdalghani Abujabal and Nora Al-Twairesh and Areeb Alowisheq and Haidar Khan},
    booktitle={The Thirteenth International Conference on Learning Representations},
    year={2025},
    url={https://openreview.net/forum?id=MscdsFVZrN}
}

@misc{Jiang2023,
  author = {Jiang, Albert and Sablayrolles, Alexandre and Mensch, Arthur and Bamford, Chris and Chaplot, Devendra Singh and de las Casas, Diego and Bressand, Florian and Lengyel, Gianna and Lample, Guillaume and Lavaud, Lélio Renard and Saulnier, Lucile and Lachaux, Marie-Anne and Stock, Pierre and Le Scao, Teven and Lavril, Thibaut and Wang, Thomas and Lacroix, Timothée and El Sayed, William},
  title = {Mistral-7B-v0.1},
  year = {2023},
  publisher = {Hugging Face},
  howpublished = {\url{https://huggingface.co/mistralai/Mistral-7B-v0.1}},
  url= {https://huggingface.co/mistralai/Mistral-7B-v0.1},
  note = {Accessed: 2025-8-1}
}

@inproceedings{Al-Laith2025,
  title = "Evaluating Calibration of {A}rabic Pre-trained Language Models on Dialectal Text",
    author = "Al-Laith, Ali  and
      Kebdani, Rachida",
    editor = "Ezzini, Saad  and
      Alami, Hamza  and
      Berrada, Ismail  and
      Benlahbib, Abdessamad  and
      El Mahdaouy, Abdelkader  and
      Lamsiyah, Salima  and
      Derrouz, Hatim  and
      Haddad Haddad, Amal  and
      Jarrar, Mustafa  and
      El-Haj, Mo  and
      Mitkov, Ruslan  and
      Rayson, Paul",
    booktitle = "Proceedings of the 4th Workshop on Arabic Corpus Linguistics (WACL-4)",
    month = jan,
    year = "2025",
    address = "Abu Dhabi, UAE",
    publisher = "Association for Computational Linguistics",
    url = "https://aclanthology.org/2025.wacl-1.8/",
    pages = "68--76"
}

@inproceedings{alharbi-etal-2025-evaluating,
    title = "Evaluating Large Language Models on Health-Related Claims Across {A}rabic Dialects",
    author = "Alharbi, Abdulsalam obaid  and
      Alsuhaibani, Abdullah  and
      Alalawi, Abdulrahman Abdullah  and
      Naseem, Usman  and
      Jameel, Shoaib  and
      Kanhere, Salil  and
      Razzak, Imran",
    editor = "El-Haj, Mo",
    booktitle = "Proceedings of the 1st Workshop on NLP for Languages Using Arabic Script",
    month = jan,
    year = "2025",
    address = "Abu Dhabi, UAE",
    publisher = "Association for Computational Linguistics",
    url = "https://aclanthology.org/2025.abjadnlp-1.11/",
    pages = "95--103"}

@article{robinson2024qasida,
  title = "{AL}-{QASIDA}: Analyzing {LLM} Quality and Accuracy Systematically in Dialectal {A}rabic",
    author = "Robinson, Nathaniel Romney  and
      Abdelmoneim, Shahd  and
      Marchisio, Kelly  and
      Ruder, Sebastian",
    editor = "Che, Wanxiang  and
      Nabende, Joyce  and
      Shutova, Ekaterina  and
      Pilehvar, Mohammad Taher",
    booktitle = "Findings of the Association for Computational Linguistics: ACL 2025",
    month = jul,
    year = "2025",
    address = "Vienna, Austria",
    publisher = "Association for Computational Linguistics",
    url = "https://aclanthology.org/2025.findings-acl.1137/",
    doi = "10.18653/v1/2025.findings-acl.1137",
    pages = "22048--22065",
    ISBN = "979-8-89176-256-5"
}

@inproceedings{Elmogtaba2024,
  title = "{LLM}-based {MT} Data Creation: Dialectal to {MSA} Translation Shared Task",
    author = "Abdelaziz, AhmedElmogtaba Abdelmoniem Ali  and
      Elneima, Ashraf Hatim  and
      Darwish, Kareem",
    editor = "Al-Khalifa, Hend  and
      Darwish, Kareem  and
      Mubarak, Hamdy  and
      Ali, Mona  and
      Elsayed, Tamer",
    booktitle = "Proceedings of the 6th Workshop on Open-Source Arabic Corpora and Processing Tools (OSACT) with Shared Tasks on Arabic LLMs Hallucination and Dialect to MSA Machine Translation @ LREC-COLING 2024",
    month = may,
    year = "2024",
    address = "Torino, Italia",
    publisher = "ELRA and ICCL",
    url = "https://aclanthology.org/2024.osact-1.14/",
    pages = "112--116"
}

@inproceedings{sajjad2020arabench,
  title = "{A}ra{B}ench: Benchmarking Dialectal {A}rabic-{E}nglish Machine Translation",
    author = "Sajjad, Hassan  and
      Abdelali, Ahmed  and
      Durrani, Nadir  and
      Dalvi, Fahim",
    editor = "Scott, Donia  and
      Bel, Nuria  and
      Zong, Chengqing",
    booktitle = "Proceedings of the 28th International Conference on Computational Linguistics",
    month = dec,
    year = "2020",
    address = "Barcelona, Spain (Online)",
    publisher = "International Committee on Computational Linguistics",
    url = "https://aclanthology.org/2020.coling-main.447/",
    doi = "10.18653/v1/2020.coling-main.447",
    pages = "5094--5107"
}

@article{alwajih2024dallah,
  title = "Dallah: A Dialect-Aware Multimodal Large Language Model for {A}rabic",
    author = "Alwajih, Fakhraddin  and
      Bhatia, Gagan  and
      Abdul-Mageed, Muhammad",
    booktitle = "Proceedings of the Second Arabic Natural Language Processing Conference",
    month = aug,
    year = "2024",
    address = "Bangkok, Thailand",
    publisher = "Association for Computational Linguistics",
    url = "https://aclanthology.org/2024.arabicnlp-1.27/",
    doi = "10.18653/v1/2024.arabicnlp-1.27",
    pages = "320--336",
}

@article{hasanaath2025arareasoner,
  title={AraReasoner: Evaluating Reasoning-Based LLMs for Arabic NLP},
  author={Ahmed Hasanaath and Aisha Alansari and Ahmed Ashraf and Chafik Salmane and Hamzah Luqman and Saad Ezzini},
  journal={arXiv preprint arXiv:2506.08768},
  year={2025}
}

@article{magdy2025jawaher,
  title = "{JAWAHER}: A Multidialectal Dataset of {A}rabic Proverbs for {LLM} Benchmarking",
    author = "Magdy, Samar Mohamed  and
      Kwon, Sang Yun  and
      Alwajih, Fakhraddin  and
      Abdelfadil, Safaa Taher  and
      Shehata, Shady  and
      Abdul-Mageed, Muhammad",
    editor = "Chiruzzo, Luis  and
      Ritter, Alan  and
      Wang, Lu",
    booktitle = "Proceedings of the 2025 Conference of the Nations of the Americas Chapter of the Association for Computational Linguistics: Human Language Technologies (Volume 1: Long Papers)",
    month = apr,
    year = "2025",
    address = "Albuquerque, New Mexico",
    publisher = "Association for Computational Linguistics",
    url = "https://aclanthology.org/2025.naacl-long.613/",
    doi = "10.18653/v1/2025.naacl-long.613",
    pages = "12320--12341",
    ISBN = "979-8-89176-189-6"
}

@article{guo2025care,
  title={Care: Aligning language models for regional cultural awareness},
  author={Geyang Guo and Tarek Naous and Hiromi Wakaki and Yukiko Nishimura and Yuki Mitsufuji and Alan Ritter and Wei Xu},
  journal={arXiv preprint arXiv:2504.05154},
  year={2025}
}

@inproceedings{alwajih2025palm,
    title = "Palm: A Culturally Inclusive and Linguistically Diverse Dataset for {A}rabic {LLM}s",
    author={Fakhraddin Alwajih and Abdellah El Mekki and Samar Mohamed Magdy and Abdelrahim A. Elmadany and Omer Nacar and El Moatez Billah Nagoudi and Reem Abdel-Salam and Hanin Atwany and Youssef Nafea and Abdulfattah Mohammed Yahya and Rahaf Alhamouri and Hamzah A. Alsayadi and Hiba Zayed and Sara Shatnawi and Serry Sibaee and Yasir Ech-Chammakhy and Walid Al-Dhabyani and Marwa Mohamed Ali and Imen Jarraya and Ahmed Oumar El-Shangiti and Aisha Alraeesi and Mohammed Anwar Al-Ghrawi and Abdulrahman S. Al-Batati and Elgizouli Mohamed and Noha Taha Elgindi and Muhammed Saeed and Houdaifa Atou and Issam Ait Yahia and Abdelhak Bouayad and Mohammed Machrouh and Amal Makouar and Dania Alkawi and Mukhtar Mohamed and Safaa Taher Abdelfadil and Amine Ziad Ounnoughene and Rouabhia Anfel and Rwaa Assi and Ahmed Sorkatti and Mohamedou Cheikh Tourad and Anis Koubaa and Ismail Berrada and Mustafa Jarrar and Shady Shehata and Muhammad Abdul-Mageed},
    booktitle = "Proceedings of the 63rd Annual Meeting of the Association for Computational Linguistics (Volume 1: Long Papers)",
    month = jul,
    year = "2025",
    address = "Vienna, Austria",
    publisher = "Association for Computational Linguistics",
    url = "https://aclanthology.org/2025.acl-long.1579/",
    doi = "10.18653/v1/2025.acl-long.1579",
    pages = "32871--32894",
    ISBN = "979-8-89176-251-0"
}

@article{alwajih2024peacock,
  title = "Peacock: A Family of {A}rabic Multimodal Large Language Models and Benchmarks",
    author = "Alwajih, Fakhraddin  and
      Nagoudi, El Moatez Billah  and
      Bhatia, Gagan  and
      Mohamed, Abdelrahman  and
      Abdul-Mageed, Muhammad",
    editor = "Ku, Lun-Wei  and
      Martins, Andre  and
      Srikumar, Vivek",
    booktitle = "Proceedings of the 62nd Annual Meeting of the Association for Computational Linguistics (Volume 1: Long Papers)",
    month = aug,
    year = "2024",
    address = "Bangkok, Thailand",
    publisher = "Association for Computational Linguistics",
    url = "https://aclanthology.org/2024.acl-long.689/",
    doi = "10.18653/v1/2024.acl-long.689",
    pages = "12753--12776"
}

@article{kadaoui2025jeem,
  title={Jeem: Vision-language understanding in four arabic dialects},
  author={Karima Kadaoui and Hanin Atwany and Hamdan Al-Ali and Abdelrahman Mohamed and Ali Mekky and Sergei Tilga and Natalia Fedorova and Ekaterina Artemova and Hanan Aldarmaki and Yova Kementchedjhieva},
  journal={arXiv preprint arXiv:2503.21910},
  year={2025}
}

@article{bhatia2024swan,
  title = "Swan and {A}rabic{MTEB}: Dialect-Aware, {A}rabic-Centric, Cross-Lingual, and Cross-Cultural Embedding Models and Benchmarks",
    author = "Bhatia, Gagan  and
      Nagoudi, El Moatez Billah  and
      El Mekki, Abdellah  and
      Alwajih, Fakhraddin  and
      Abdul-Mageed, Muhammad",
    editor = "Chiruzzo, Luis  and
      Ritter, Alan  and
      Wang, Lu",
    booktitle = "Findings of the Association for Computational Linguistics: NAACL 2025",
    month = apr,
    year = "2025",
    address = "Albuquerque, New Mexico",
    publisher = "Association for Computational Linguistics",
    url = "https://aclanthology.org/2025.findings-naacl.263/",
    doi = "10.18653/v1/2025.findings-naacl.263",
    pages = "4654--4670",
    ISBN = "979-8-89176-195-7"
}

@article{ayash2025saudiculture,
  title={SaudiCulture: A benchmark for evaluating large language models’ cultural competence within Saudi Arabia},
  author={Lama Ayash and Hassan Alhuzali and Ashwag Alasmari and Sultan Aloufi},
  journal={Journal of King Saud University Computer and Information Sciences},
  volume={37},
  number={6},
  pages={123},
  year={2025},
  publisher={Springer},
  doi       = {10.1007/s44443-025-00137-9},
  url       = {https://doi.org/10.1007/s44443-025-00137-9}
}
\clearpage
\appendix
\section*{Appendices}
\addcontentsline{toc}{section}{Appendices}

\section{Prompt Example} \label{sec:prompt}

\begin{figure}[h!]
    \centering
    \includegraphics[width=\textwidth]{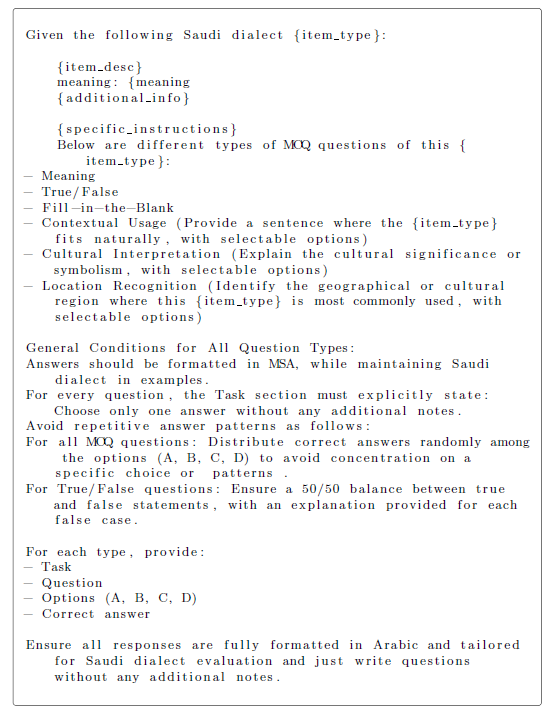} 
\end{figure}






\clearpage


\section{Annotation Guideline} \label{sec:annotation_guideline}
\vspace{0.5em}
\begin{center}
\noindent\vspace{0.5em}

\includegraphics[width=0.9\textwidth]{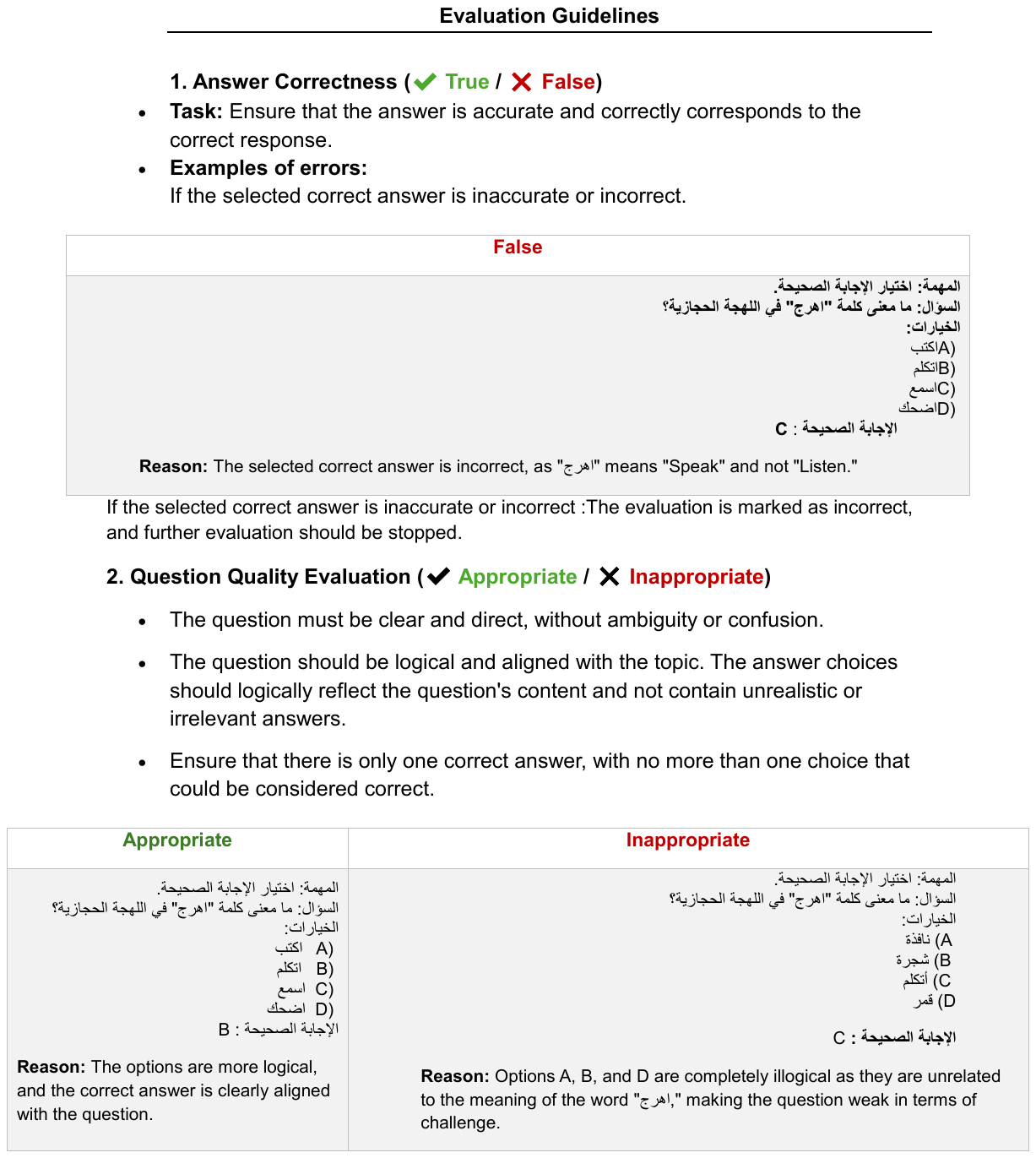}
\end{center}

\section{The Annotation interface} \label{sec:interface}
\vspace{1em}
\begin{center}
    \includegraphics[width=\textwidth]{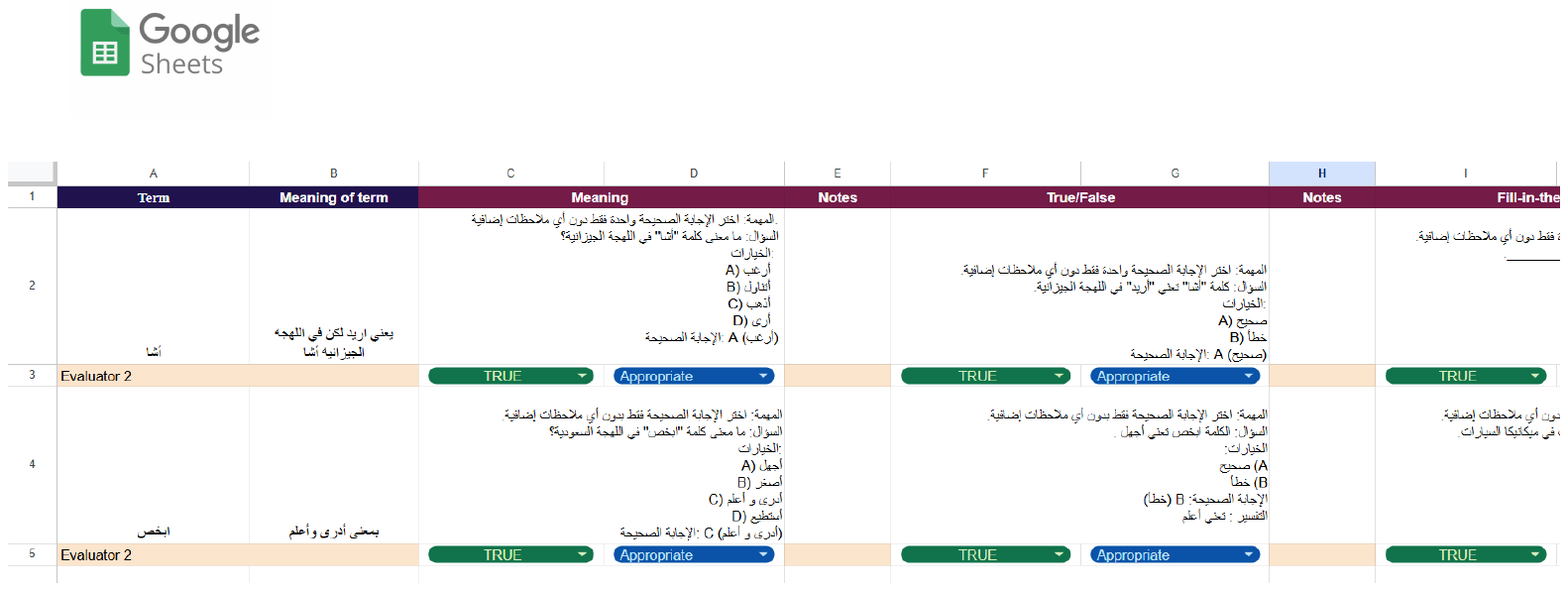}
    
    \begin{minipage}{0.9\textwidth}
        \centering
    \end{minipage}
\end{center}



\vspace{7em}
\section{Detailed Performance} \label{sec:Detailed-Performance}
\noindent\vspace{0.5em}

\begin{center}
{\footnotesize 
\renewcommand{\arraystretch}{1.4}
\setlength{\tabcolsep}{4pt}
\onecolumn
\begin{longtable}{lllcccc}
\caption{Detailed evaluation of the results across content and question types, with the best results highlighted in dark green to represent the highest scores.} \\
\toprule
\textbf{Model} & \textbf{Content Type} & \textbf{Question Type} & \textbf{Accuracy} & \textbf{Precision} & \textbf{Recall} & \textbf{F1-score} \\
\midrule
\endfirsthead

\toprule
\textbf{Model} & \textbf{Content Type} & \textbf{Question Type} & \textbf{Accuracy} & \textbf{Precision} & \textbf{Recall} & \textbf{F1-score} \\
\midrule
\endhead

\midrule \multicolumn{7}{r}{{Continued on next page}} \\
\endfoot

\bottomrule

\endlastfoot

\multirow{18}{*}{\includegraphics[width=0.7cm]{ALLaM1.png} \textbf{ALLaM}} 
& \multirow{6}{*}{Word} 
    & Meaning & 24.42\% & 28.08\% & 25.32\% & 10.48\% \\
&   & True/False & 21.88\% & 20.45\% & 24.27\% & 9.21\% \\
&   & Fill-in-the-Blank &\cellcolor{green!15} 44.24\% & 39.54\% & 32.19\% & 28.19\% \\
    &   & Contextual Usage &\cellcolor{green!15} 42.08\% & 48.53\% & 27.49\% & 19.46\% \\
&   & Cultural Interpretation &\cellcolor{green!15} 42.33\% & 38.99\% & 25.42\% & 15.86\% \\
&   & Location Recognition & \cellcolor{green!50}67.44\% & 27.63\% & 25.04\% & 20.46\% \\
\cmidrule(l){2-7}
& \multirow{6}{*}{Phrase} 
    & Meaning & \cellcolor{green!15} 46.03\% & 48.09\% & 35.48\% & 24.49\% \\
&   & True/False & 22.00\% & 5.55\% & 24.26\% & 9.03\% \\
&   & Fill-in-the-Blank &\cellcolor{green!15} 43.53\% & 32.71\% & 27.84\% & 23.35\% \\
&   & Contextual Usage & 37.26\% & 50.58\% & 23.49\% & 17.40\% \\
&   & Cultural Interpretation & 39.18\% & 9.82\% & 24.86\% & 14.07\% \\
&   & Location Recognition & \cellcolor{green!50}65.50\% & 16.48\% & 24.92\% & 19.84\% \\
\cmidrule(l){2-7}
& \multirow{6}{*}{Proverb} 
    & Meaning & \cellcolor{green!15} 46.15\% & 35.00\% & 32.14\% & 25.40\% \\
&   & True/False & 30.77\% & 29.17\% & 32.14\% & 29.09\% \\
&   & Fill-in-the-Blank &\cellcolor{green!30} 55.56\% & 38.51\% & 26.15\% & 21.53\% \\
&   & Contextual Usage &\cellcolor{green!15} 45.00\% & 47.22\% & 51.30\% & 42.04\% \\
&   & Cultural Interpretation & \cellcolor{green!50}60.00\% & 58.79\% & 52.46\% & 52.99\% \\
&   & Location Recognition & \cellcolor{green!30} 53.25\% & 31.00\% & 30.36\% & 27.73\% \\
\midrule
\multirow{18}{*}{\includegraphics[width=0.8cm]{lama.jpg} \textbf{LLaMA}} 
& \multirow{6}{*}{Word} 
    & Meaning & 25.98\% & 49.80\% & 25.49\% & 13.15\% \\
&   & True/False & 27.24\% & 25.12\% & 24.93\% & 13.16\% \\
&   & Fill-in-the-Blank &\cellcolor{green!15} 42.64\% & 38.67\% & 27.45\% & 20.38\% \\
&   & Contextual Usage &\cellcolor{green!15} 43.55\% & 44.36\% & 22.98\% & 18.27\% \\
&   & Cultural Interpretation & \cellcolor{green!15} 41.99\% & 28.15\% & 20.12\% & 13.14\%  \\
&   & Location Recognition & \cellcolor{green!50}66.32\% & 30.85\% & 21.87\% & 20.66\% \\
\cmidrule(l){2-7}
& \multirow{6}{*}{Phrase} 
    & Meaning &\cellcolor{green!15} 42.86\% & 23.33\% & 24.88\% & 16.46\% \\
&   & True/False & 29.56\% & 28.41\% & 26.60\% & 14.20\% \\
&   & Fill-in-the-Blank &\cellcolor{green!15} 45.91\% & 31.11\% & 27.16\% & 20.46\% \\
&   & Contextual Usage & 38.33\% & 37.73\% & 23.43\% & 18.02\% \\
&   & Cultural Interpretation &\cellcolor{green!15} 40.69\% & 42.47\% & 26.00\% & 16.90\% \\
&   & Location Recognition & \cellcolor{green!50}63.32\% & 24.41\% & 21.79\% & 20.35\% \\
\cmidrule(l){2-7}
& \multirow{6}{*}{Proverb} 
    & Meaning & 30.77\% & 10.26\% & 33.33\% & 15.69\% \\
&   & True/False & 32.50\% & 41.31\% & 38.33\% & 21.57\% \\
&   & Fill-in-the-Blank &\cellcolor{green!30} 56.79\% & 38.82\% & 28.12\% & 23.35\% \\
&   & Contextual Usage &\cellcolor{green!15} 40.00\% & 34.74\% & 26.56\% & 16.96\% \\
&   & Cultural Interpretation &\cellcolor{green!15} 46.25\% & 24.04\% & 25.81\% & 17.30\% \\
&   & Location Recognition &\cellcolor{green!30} 54.55\% & 13.64\% & 25.00\% & 17.65\% \\
\midrule 
\multirow{18}{*}{\includegraphics[width=0.8cm]{jais.png} \textbf{Jais}} 
& \multirow{6}{*}{Word} 
    & Meaning & 23.89\% & 30.87\% & 25.26\% & 10.02\% \\
&   & True/False & 22.26\% & 22.82\% & 24.79\% & 9.26\% \\
&   & Fill-in-the-Blank & 39.57\% & 35.41\% & 20.48\% & 12.29\% \\
&   & Contextual Usage &\cellcolor{green!15} 40.36\% & 33.62\% & 25.55\% & 15.98\% \\
&   & Cultural Interpretation &\cellcolor{green!15} 44.31\% & 52.65\% & 27.68\% & 20.96\% \\
&   & Location Recognition & \cellcolor{green!50}67.57\% & 16.91\% & 24.97\% & 20.17\% \\
\cmidrule(l){2-7}
& \multirow{6}{*}{Phrase} 
    & Meaning &\cellcolor{green!15} 42.86\% & 14.29\% & 33.33\% & 20.00\% \\
&   & True/False & 23.11\% & 40.92\% & 33.53\% & 12.75\% \\
&   & Fill-in-the-Blank &\cellcolor{green!15} 41.59\% & 51.99\% & 25.94\% & 16.68\% \\
&   & Contextual Usage & 35.76\% & 31.09\% & 20.46\% & 11.77\% \\
&   & Cultural Interpretation & 39.61\% & 45.27\% & 25.39\% & 16.10\% \\
&   & Location Recognition & \cellcolor{green!50}65.72\% & 41.48\% & 25.15\% & 20.30\% \\
\cmidrule(l){2-7}
& \multirow{6}{*}{Proverb} 
    & Meaning & 30.77\% & 10.26\% & 33.33\% & 15.69\% \\
&   & True/False & 21.25\% & 7.08\% & 33.33\% & 11.68\% \\
&   & Fill-in-the-Blank &\cellcolor{green!30} 51.85\% & 12.96\% & 25.00\% & 17.07\% \\
&   & Contextual Usage & 37.50\% & 9.38\% & 25.00\% & 13.64\% \\
&   & Cultural Interpretation &\cellcolor{green!15} 45.00\% & 11.25\% & 25.00\% & 15.52\% \\
&   & Location Recognition &\cellcolor{green!30} 54.55\% & 13.64\% & 25.00\% & 17.65\% \\

\midrule

\multirow{18}{*}{\includegraphics[width=0.5cm]{M-logo.png} \textbf{Mistral}} 
& \multirow{6}{*}{Word} 
    & Meaning & 32.08\% & 29.91\% & 26.60\% & 22.09\% \\
&   & True/False & \cellcolor{green!30}58.40\% & 58.15\% & 59.70\% & 56.56\% \\
&   & Fill-in-the-Blank &\cellcolor{green!15} 42.97\% & 10.74\% & 25.00\% & 15.03\% \\
&   & Contextual Usage & 39.48\% & 9.87\% & 25.00\% & 14.15\% \\
&   & Cultural Interpretation &\cellcolor{green!15} 45.07\% & 11.27\% & 25.00\% & 15.53\% \\
&   & Location Recognition &\cellcolor{green!50} 69.77\% & 17.44\% & 25.00\% & 20.55\% \\
\cmidrule(l){2-7}
& \multirow{6}{*}{Phrase} 
    & Meaning & 38.71\% & 27.44\% & 26.34\% & 24.64\% \\
&   & True/False & \cellcolor{green!30}56.00\% & 58.28\% & 59.38\% & 55.36\% \\
&   & Fill-in-the-Blank &\cellcolor{green!15} 41.38\% & 35.31\% & 25.11\% & 14.82\% \\
&   & Contextual Usage & 35.12\% & 7.04\% & 20.00\% & 10.41\% \\
&   & Cultural Interpretation & 39.39\% &9.85\% & 25.00\% & 14.13\% \\
&   & Location Recognition & \cellcolor{green!30}58.70\% & 30.77\% & 46.51\% & 34.36\% \\
\cmidrule(l){2-7}
& \multirow{6}{*}{Proverb} 
    & Meaning & 30.77\% & 11.11\% & 33.33\% & 16.67\% \\
&   & True/False & \cellcolor{green!15}46.15\% & 30.00\% & 33.33\% & 31.58\% \\
&   & Fill-in-the-Blank &\cellcolor{green!30} 50.62\% & 12.97\% & 24.40\% & 16.94\% \\
&   & Contextual Usage & 36.25\% & 9.18\%  &24.17\%  & 13.30\% \\
&   & Cultural Interpretation & \cellcolor{green!15}45.00\% & 11.25\% &25.00\%  & 15.52\% \\
&   & Location Recognition &\cellcolor{green!30} 53.85\% & 35.42\% & 35.00\% & 34.85\% \\
\midrule
\multirow{18}{*}{\includegraphics[width=0.7cm]{Qwen.png} \textbf{Qwen}} 
& \multirow{6}{*}{Word} 
    & Meaning &\cellcolor{green!50} 64.47\% & 35.52\% & 37.59\% & 35.81\% \\
&   & True/False &\cellcolor{green!50} 75.13\% & 31.78\% & 30.60\% & 30.82\% \\
&   & Fill-in-the-Blank &\cellcolor{green!50} 60.51\% & 47.27\% & 54.62\% & 48.08\% \\
&   & Contextual Usage &\cellcolor{green!50} 64.67\% & 52.41\% & 54.08\% & 52.96\% \\
&   & Cultural Interpretation &\cellcolor{green!50} 63.74\% & 40.79\% & 44.49\% & 41.28\% \\
&   & Location Recognition&\cellcolor{green!30} 51.45\% & 44.65\% & 56.56\% & 43.11\% \\
\cmidrule(l){2-7}
& \multirow{6}{*}{Phrase} 
    & Meaning &\cellcolor{green!50} 61.29\% & 36.27\% & 37.65\% & 36.12\% \\
&   & True/False & 18.49\% & 32.23\% & 31.89\% & 11.45\% \\
&   & Fill-in-the-Blank & 33.33\% & 25.04\% & 27.41\% & 16.58\% \\
&   & Contextual Usage & 33.75\% & 24.97\% & 16.79\% & 13.41\% \\
&   & Cultural Interpretation &\cellcolor{green!50} 63.93\% & 39.62\% & 46.12\% & 40.16\% \\
&   & Location Recognition &\cellcolor{green!30} 51.97\% & 42.94\% & 51.74\% & 40.61\% \\
\cmidrule(l){2-7}
& \multirow{6}{*}{Proverb} 
    & Meaning &\cellcolor{green!15} 47.56\% & 24.38\% & 12.04\% & 16.12\% \\
&   & True/False & 31.71\% & 23.13\% & 33.11\% & 13.64\% \\
&   & Fill-in-the-Blank & 31.71\% & 20.00\% & 14.84\% & 15.69\% \\
&   & Contextual Usage &\cellcolor{green!50} 67.50\% & 59.91\% & 59.74\% & 57.64\% \\
&   & Cultural Interpretation &\cellcolor{green!30} 50.00\% & 24.40\% & 12.65\% & 16.67\% \\
&   & Location Recognition & 31.71\% & 24.07\% & 8.02\% & 12.04\% \\
\midrule

\multirow{18}{*}{\includegraphics[width=0.6cm]{ACE.png} \textbf{ACEGPT}} 
& \multirow{6}{*}{Word} 
    & Meaning &\cellcolor{green!15} 43.80\% & 26.37\% & 22.11\% & 21.07\% \\
&   & True/False & 37.22\% & 19.86\% & 20.11\% & 14.96\% \\
&   & Fill-in-the-Blank &\cellcolor{green!15}41.84\% & 35.63\% & 41.09\% & 33.47\% \\
&   & Contextual Usage &\cellcolor{green!30} 54.53\% & 47.50\% & 45.84\% & 44.14\% \\
&   & Cultural Interpretation &\cellcolor{green!15} 40.23\% & 33.23\% & 31.89\% & 26.85\% \\
&   & Location Recognition &\cellcolor{green!15} 40.28\% & 36.79\% & 38.93\% & 29.34\% \\
\cmidrule(l){2-7}
& \multirow{6}{*}{Phrase} 
    & Meaning & 39.47\% & 23.94\% & 22.26\% & 21.26\% \\
&   & True/False &\cellcolor{green!15} 40.48\% & 27.03\% & 27.35\% & 20.72\% \\
&   & Fill-in-the-Blank & 36.84\% & 34.81\% & 41.95\% & 29.54\% \\
&   & Contextual Usage &\cellcolor{green!15} 49.12\% & 48.07\% & 48.73\% & 41.73\% \\
&   & Cultural Interpretation & 39.29\% & 35.71\% & 33.49\% & 26.63\% \\
&   & Location Recognition & 36.95\% & 34.43\% & 41.52\% & 26.72\% \\
\cmidrule(l){2-7}
& \multirow{6}{*}{Proverb} 
    & Meaning &\cellcolor{green!15} 42.31\% & 44.70\% & 32.44\% & 34.03\% \\
&   & True/False & 23.81\% & 41.67\% & 43.38\% & 23.64\% \\
&   & Fill-in-the-Blank & 28.57\% & 26.67\% & 15.56\% & 19.64\% \\
&   & Contextual Usage &\cellcolor{green!15} 43.24\% & 50.75\% & 39.73\% & 32.26\% \\
&   & Cultural Interpretation & 28.26\% & 49.38\% & 38.98\% & 29.33\% \\
&   & Location Recognition & 38.67\% & 30.48\% & 37.80\% & 28.22\% \\
\end{longtable}
}
\end{center}


\end{document}